\PassOptionsToPackage{table}{xcolor}

\documentclass[research]{fcs}

\usepackage[utf8]{inputenc} 
\usepackage[T1]{fontenc}    
\usepackage{hyperref}
\usepackage{url} 
\usepackage{booktabs}
\usepackage{cite}
\usepackage{graphicx}
\usepackage{amsmath}
\usepackage{amsfonts}
\usepackage{amsthm}
\usepackage{amssymb}
\usepackage{algorithm}
\usepackage{algorithmic}
\usepackage{enumerate}
\usepackage{subcaption}

\usepackage{multirow}
\usepackage{pdfpages}

\usepackage{array}
\usepackage{textcomp}
\usepackage{verbatim}
\usepackage{balance}

\newcommand{\tohide}[1]{} 

\newcommand{\ours}[1]{SPA++}
\newcommand{\ourloss}[1]{SPA-Loss}
\newcommand{\vpara}[1]{\vspace{0.05in}\noindent\textbf{#1 }}
\newcommand{\vpait}[1]{\vspace{0.05in}\noindent\textit{#1 }}
\newcommand{\transpose}[1]{\ensuremath{#1^{\scriptscriptstyle T}}}

\definecolor{Gray}{gray}{0.9}

\title{
\ours{}: Generalized Graph Spectral Alignment for Versatile Domain Adaptation
}
\author[1]{Zhiqing XIAO}
\author[2]{Haobo WANG}
\author[3]{Xu LU}
\author[1]{Wentao YE}
\author[1]{Gang CHEN}
\author*[1]{Junbo ZHAO}
\address[1]{College of Computer Science and Technology, Zhejiang University, Zhejiang 310027, China}
\address[2]{School of Software Technology, Zhejiang University, Zhejiang 315048, China}
\address[3]{College of Computer Science and Electronic Engineering, Hunan University, Hunan 410082, China}
\fcssetup{
  received       = {Mar 25, 2025},
  accepted       = {Jul 31, 2025},
  corr-email     = {j.zhao@zju.edu.cn},
}

\begin{abstract}
Domain Adaptation (DA) aims to transfer knowledge from a labeled source domain to an unlabeled or sparsely labeled target domain under domain shifts. Most prior works focus on capturing the inter-domain transferability but largely overlook rich intra-domain structures, which empirically results in even worse discriminability. To tackle this tradeoff, we propose a generalized graph SPectral Alignment framework, SPA++. Its core is briefly condensed as follows: (1)-by casting the DA problem to graph primitives, it composes a coarse graph alignment mechanism with a novel spectral regularizer toward aligning the domain graphs in eigenspaces; (2)-we further develop a fine-grained neighbor-aware propagation mechanism for enhanced discriminability in the target domain; (3)-by incorporating data augmentation and consistency regularization, SPA++ can adapt to complex scenarios including most DA settings and even challenging distribution scenarios. Furthermore, we also provide theoretical analysis to support our method, including the generalization bound of graph-based DA and the role of spectral alignment and smoothing consistency. Extensive experiments on benchmark datasets demonstrate that SPA++ consistently outperforms existing cutting-edge methods, achieving superior robustness and adaptability across various challenging adaptation scenarios.
\end{abstract}


\begin{document}

\section{Introduction}
Domain adaptation (DA) is a long-standing and widely studied problem in the field of computer vision \cite{gong2012geodesic, ganin2016domain, tzeng2017adversarial, saito2020universal, lee2019drop}. It aims to transfer knowledge from label-rich source domains to label-scarce target domains, under the presence of dataset shifts \cite{joaquin2008dataset} or domain shifts \cite{tommasi2016learning}.
Most existing DA methods aim to learn domain-invariant feature representations, largely inspired by theoretical analysis from domain adaptation theory \cite{bendavid2006analysis}. Generally, these methods fall into two main categories: moment matching methods \cite{li2020enhanced, long2015learning, sun2016deep}, and adversarial learning methods \cite{ganin2015unsupervised, saito2018maximum, ganin2016domain, long2018conditional}.

A core challenge is achieving a proper balance between inter-domain transferability and intra-domain discriminability. In other words, a successful DA model should align features across domains effectively, while preserving the structure necessary for accurate classification within each domain. Adversarial methods try to implicitly bridge the domain gap by enforcing feature indistinguishability through domain confusion \cite{cui2020gradually, li2021bi, chen2022reusing}. However, such approaches often suffer from a degradation in discriminability, as demonstrated in recent studies \cite{chen2019transferability, kundu2022balancing}.

To mitigate this issue, a promising line of work introduces graph-based algorithms \cite{chen2020graph, zhang2018structural, luo2022attention}, which leverage rich topological structures to model intra-domain correlations. 
This line of methods typically constructs self-correlation graphs, encouraging homophily—nearby nodes tend to share labels \cite{mcpherson2001birds}. However, transferring knowledge across domains with such intra-domain graphs remains challenging. 
Existing methods often rely on explicit graph matching, requiring rigid point-wise node alignment across domains, which is both computationally expensive and unnecessarily strict for DA tasks \cite{chen2020graph, zhang2018structural, luo2022attention}.

\begin{figure*}[t]
\centering
\includegraphics[width=\textwidth]{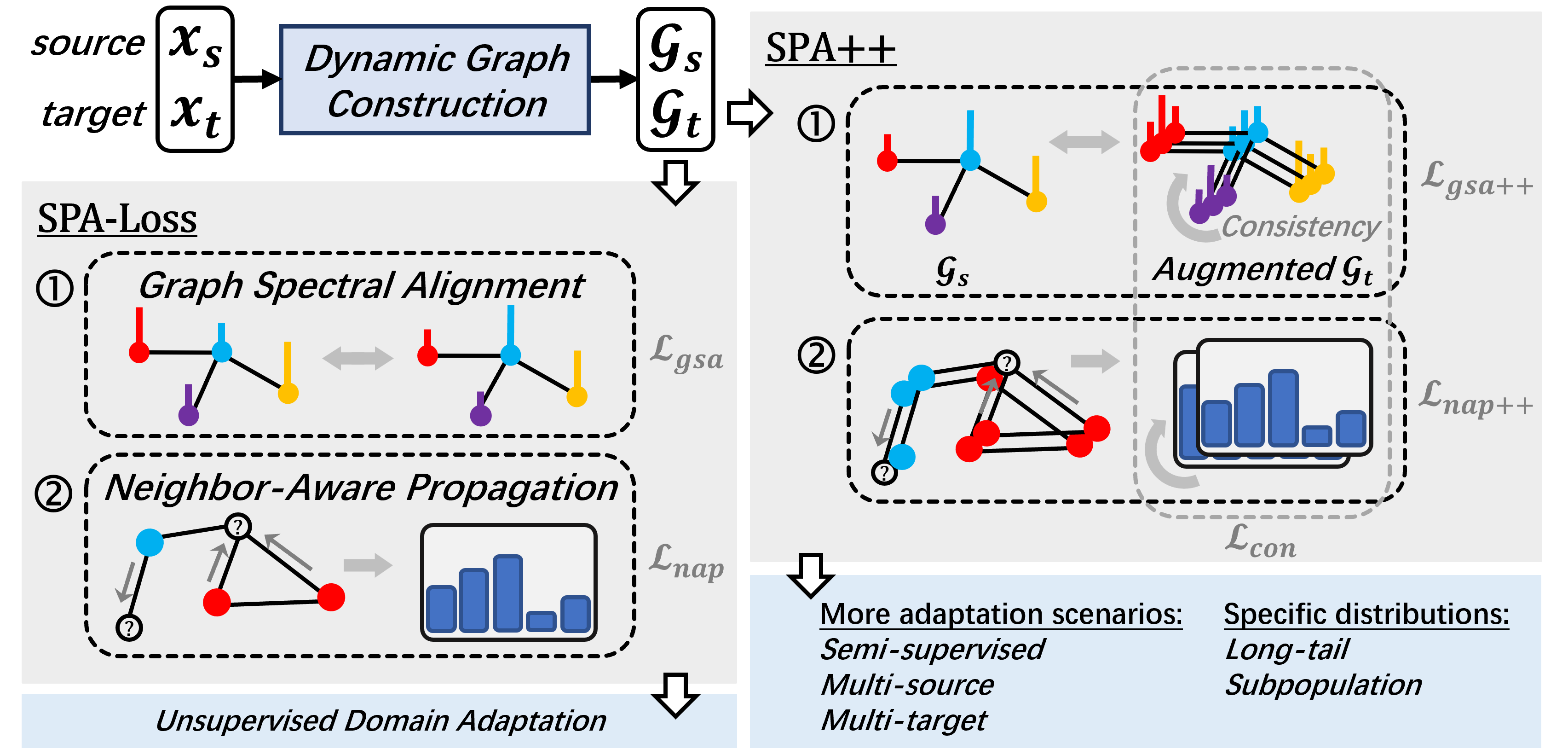}
\caption{
The overall architecture of \ourloss{} and \ours{}.
The final objective of \ourloss{} mainly depends on  
neighbor-aware propagation loss $\mathcal{L}_{nap}$, and 
graph spectral alignment loss $\mathcal{L}_{spa}$.
And the \ours{} further generalizes \ourloss{} with augmentation and consistency, which allows the whole model to adapt to more scenarios and even specific adaptation challenges. 
}
\label{fig:framework}
\end{figure*}

In this work, we propose a novel graph spectral alignment framework for DA, designed to jointly leverage balancing inter-domain transferability and intra-domain discriminability hierarchically and flexibly.
As illustrated in Figure \ref{fig:framework}, 
our method hierarchically decomposes the alignment problem into a coarse-grained graph alignment and a fine-grained propagation module.
At the core of our method lies a spectral regularizer, which projects source and target domain graphs into their respective eigenspaces and aligns them based on their spectral properties. 
This enables effective coarse-grained structure transfer between domains without relying on point-wise node matching. Subsequently, a neighbor-aware propagation mechanism performs fine-grained message propagation by leveraging predictions from confident neighbors.
This strategy refines the transferred structure and enhances discriminability.
We incorporate data augmentation and progressive consistency strategies into the adaptation process to further improve generalization and robustness as \ours{}.
Going a step further, \ours{} extends our base model to a broader range of DA scenarios, including semi-supervised DA, multi-domain DA, and even more complex distribution challenges, showcasing strong adaptability across versatile real-world scenarios. 

This journal version of \ours{} expands our conference paper \cite{xiao2024spa} along several principal axes: 
(i) It adds a consistency regularization component to stabilize the spectral alignment and neighbor propagation modules.
(ii) It generalizes the framework to more challenging adaptation scenarios, including SSDA, MSDA, MTDA, long-tail, and sub-population shifts.
(iii) It contributes a formal theoretical analysis to graph spectral alignment with smoothing consistency.
The main extensions are detailed below:
\begin{itemize}
    \item {
    \textit{Setup:} 
    Our conference version focuses on the UDA setting but fails to handle SSDA, MSDA, and MTDA, not to mention the more challenging settings such as long-tail and subpopulation distribution. 
    This journal version expands the UDA settings to more complex scenarios by introducing augmentation and processing consistency, leading to more robust and consistently high performance.
    Overall, this version covers the mainstream DA settings, demonstrating the broad application of our approaches.}
    \item {
    \textit{Method:}
    This journal version improves \ourloss{} by stabilizing the estimation of graph alignment and neighbor propagation with a consistency regularization under auto-augmentations.
    Specifically, we minimize the L2 distance between the predictions of augmentation graphs.
    We integrate them into the overall pipeline as \ours{}.
    Experimental results show that consistency regularization enables a more accurate estimation and prediction. Further comparison with prior works shows that the proposed pipeline outperforms the conventional one in DA tasks. 
    }
    \item {
    \textit{Theory:}
    \ourloss{} in our previous conference paper introduced a novel graph spectral alignment perspective for UDA from an insightful viewpoint, but it lacked a formal theoretical analysis.
    In this journal version, we provide theoretical foundations on the generalization bound of graph-based DA, the principles behind graph spectral alignment, and the role of smoothing consistency, thereby complementing and strengthening the theoretical depth of our earlier work.
    }
    \item {
    \textit{Experiment:}
    Our conference version empirically evaluates \ourloss{} mainly on UDA settings, but this journal version evaluates more challenging data distribution scenarios.
    Extensive ablation studies confirm the individual contributions of our spectral alignment module, propagation mechanism, and the generalized design of \ours{}.
    Comprehensive model analysis further demonstrates the superiority of \ours{} in efficacy, robustness, discriminability, and transferability across versatile DA settings. 
    }
\end{itemize}

\section{Related Work}

\subsection{Domain Adaptation}
DA addresses domain shifts or dataset discrepancies challenges \cite{joaquin2008dataset, tommasi2016learning}, which has been extensively studied in the computer vision field \cite{ganin2015unsupervised, gong2012geodesic, tzeng2017adversarial, long2018conditional, saito2020universal, lee2019drop, li2020enhanced, sun2016deep, long2015learning}.

By the availability of labeled data in the target domain, DA can be categorized as 
\textit{Unsupervised} and \textit{Semi-supervised} DA settings:
\begin{itemize}
    \item {
    UDA assumes no labeled data in the target domain. Most UDA methods focus on learning domain-invariant features to align source and target feature distributions \cite{saito2018maximum, li2020maximum}.
    Classical methods include moment-matching approaches \cite{long2015learning, sun2016deep, li2020enhanced} that explicitly minimize distribution discrepancies, and adversarial learning methods \cite{ganin2015unsupervised, tzeng2017adversarial, ganin2016domain, long2018conditional} that use a domain classifier for domain-invariant feature learning.
    }
    \item {
    SSDA assumes only a small amount of labeled data is available in the target domain. This setting seeks to leverage both the labeled and unlabeled data in the target domain to improve adaptation performance.
    UDA methods are very effective at aligning feature distributions of source and target domains without any target supervision, but perform poorly when even a few labeled examples are available in the target domain. 
    SSDA methods address this limitation by combining UDA strategies with supervised learning on the labeled target data \cite{ganin2015unsupervised, cui2020towards, rahman2023semi}.
    }
\end{itemize}

By the domain setup, DA can further be categorized as \textit{Multi-source} and \textit{Multi-target} DA settings:
\begin{itemize}
    \item {
    MSDA aims to transfer knowledge from multiple labeled source domains to a single unlabeled target domain.
    Some MSDA methods extend classic UDA methods, such as moment matching or adversarial learning strategies, to align multiple source domains with the target domain \cite{xu2018deep, peng2019moment, zhao2018adversarial, kang2020contrastive}.
    }
    \item {
    MTDA from a single labeled source domain to multiple unlabeled target domains. It often involves modeling relationships between target domains to improve generalization.
    It usually needs to model the relationships between target domains to improve generalization \cite{french2017self, peng2019domain, nguyen2021unsupervised, isobe2021multi, kumar2023conmix}.
    }
\end{itemize}

By the data distribution in the target domain, DA includes the following specific challenges:
\begin{itemize}
    \item {
    \textit{Long-tail} addresses imbalanced data distributions, where a few categories dominate the data while others appear infrequently. Methods in this setting tackle both domain shifts and long-tail challenges \cite{TALLY, yang2022multi, li2021imbalanced, shi2022pairwise}.
    }
    \item {
    \textit{Subpopulation} focuses on adapting to specific subsets of data within the target domain, which may exhibit unique distributions or fine-grained differences compared to the overall target domain \cite{yang2022multi, haochen2022beyond, garg2023rlsbench}.
    }
\end{itemize}

Our proposed method, \ours{}, primarily focuses on unsupervised domain adaptation but can also be extended to semi-supervised, multi-domain settings and even with typical data distribution cases. By leveraging graph spectral alignment, \ours{} achieves state-of-the-art results across versatile DA scenarios.

\subsection{Graph Data Mining}
Graphs are widely used to model pairwise relationships in various domains, such as biology \cite{yi2022graph}, social networks \cite{li2022distilling}, and finance \cite{song2023research}. 
Graph data mining has been a longstanding research focus \cite{bruna2014spectral, kipf2017semi, velickovic2018graph}, with applications in computer vision tasks like image retrieval \cite{karpathy2015deep}, object detection \cite{li2022sigma}, and image classification \cite{vasudevan2023image}.
Classical Graph Neural Networks (GNNs) were first introduced by Bruna et al. \cite{bruna2014spectral}, leveraging spectral graph theory to perform convolution operations,
and subsequent methods improved computational efficiency and scalability.
ChebNet \cite{defferrard2016convolutional} uses Chebyshev polynomials to approximate graph filters, and GCN \cite{kipf2017semi} simplifies graph convolutions using local neighborhoods.
Recently, graph-based domain adaptation has gained attention due to the increasing availability of graph-structured data. 
Graph DA methods extend traditional DA techniques to graph data by aligning graph topologies, node features, and label distributions across domains. 
Representative works include \cite{cai2023graph} that aligns node embeddings across domains, \cite{pang2024sagda} that leverages adversarial learning to reduce graph distribution discrepancies, \cite{you2023graph} that introduces frequency algorithms to enhance the transferability of GNNs across domains,
and the very recent \cite{yang2025dgsda} that disentangles attribute- and topology-level alignment and employs a Bernstein-polynomial approximation to flexibly match the spectral filters of the source and target graphs.
Besides, some image domain adaptation methods also leverage Laplace graphs. For instance, \cite{luo2022attention} introduces Laplace graphs with an attention mechanism to preserve manifold structure and emphasize class similarity.

Our proposed methods utilize the good properties of simple graphs from spectral graph theory.
For instance, algebraic connectivity, \emph{i.e.}, the second-smallest non-zero Laplacian eigenvalue, reflects graph connectivity \cite{chung1997spectral}.
Insight by this, our \ourloss{} and \ours{} distinguish themselves by utilizing graph spectral alignment via leveraging the eigenvalues of the graph Laplacian matrix, providing a novel perspective for domain adaptation.

\subsection{Self-training}
Self-training is a semi-supervised learning method that enhances supervised models by generating pseudo-labels for unlabeled data based on model predictions, which has been explored in lots of works \cite{iscen2019label, lee2013pseudo, liu2019learning, wu2018unsupervised, xie2021n, litrico2023guiding}.
These methods pick up the class with the maximum predicted probability as true labels each time the weights are updated.
Filtering strategies and iterative approaches are employed to improve the quality of pseudo labels.
Some works also enforce the model to produce consistent predictions for unlabeled data under different perturbations, such as data augmentations or adversarial noise \cite{sosea2022leveraging, zhang2023refined}. 
To effectively utilize the unlabeled target domain data, the self-training technique has also been applied to some domain adaptation methods \cite{liang2021domain, zhang2020progressive, wang2021uncertainty, kundu2022subsidiary}.
This kind of method proves beneficial when the labeled data is limited.
However,  it relies on an assumption of the unlabeled data following the same distribution as the labeled data and requires accurate initial model predictions for precise results \cite{pham2021meta, arazo2020pseudo}. 

Our intra-domain alignment helps the neighbor-aware propagation mechanism generate more precise labels.
By incorporating intra-domain alignment and neighbor-aware propagation, our \ourloss{} and \ours{} improve pseudo-label precision, achieving impressive results in a series of DA scenarios.
\section{Preliminaries} \label{sec:preliminaries}

\subsection{Problem Definition}
    Our method can extend to versatile DA scenarios, while here, for simplicity, we only introduce the preliminaries, the proposed method, and theoretical analysis under the setting of UDA.
    Below is the detailed definition of the UDA setting.
    Given source domain data $\mathcal{D}_s=\left\{\left(x_i^s, y_i^s\right)\right\}_{i=1}^{N_s}$
    of $N_s$ labeled samples associated with $C_s$ categories from $\mathcal{X}_s \times \mathcal{Y}_s$ and target domain data $\mathcal{D}_t=\left\{x_i^t\right\}_{i=1}^{N_t}$ of $N_t$ unlabeled samples associated with $C_t$ categories from $\mathcal{X}_t$.
    We assume that the domains share the same feature and label space but follow different marginal data distributions, following the covariate shift \cite{hidetoshi2000improving}; 
    that is, $P\left(\mathcal{X}_s\right) \neq P\left(\mathcal{X}_t\right)$ 
    but $P\left(\mathcal{Y}_s \mid \mathcal{X}_s\right) = P\left(\mathcal{Y}_t \mid \mathcal{X}_t\right)$. 
    DA just occurs when the underlying distributions corresponding to the source and target domains in the shared label space are different but similar enough to make sense of the transfer \cite{david2010impossibility}. 
    The goal of unsupervised domain adaptation is to predict the label $\left\{y_i^t\right\}_{i=1}^{N_t}$ in the target domain, where $y_i^t \in \mathcal{Y}_t$, and the source task is $\mathcal{X}_s \rightarrow \mathcal{Y}_s$ assumed to be the same as the target task $\mathcal{X}_t \rightarrow \mathcal{Y}_t$.

\subsection{Adversarial Domain Adaptation}
    The existing adversarial domain adaptation methods, \emph{e.g.}, Domain Adversarial Neural Network (DANN) \cite{ganin2015unsupervised} have become significantly influential in the DA field. 
    The basic idea is to learn transferable features that explicitly reduce the domain shift. 
    Similar to standard supervised classification methods, these approaches include a feature extractor 
    $F(\cdot)$
    and a category classifier 
    $C(\cdot)$.
    Additionally, 
    a domain classifier 
    $D(\cdot)$ is proposed as the adversary to play a two-player minimax game against the feature extractor. 
    The feature extractor $F(\cdot)$ can be designed to learn source features and target features from source domain samples and target domain samples,     
    \emph{i.e.}, 
    $\mathbf{f}_s = F( \mathbf{x}_s )$ and
    $\mathbf{f}_t = F( \mathbf{x}_t )$.
    The supervised classification loss $\mathcal{L}_{cls}$ and 
    domain adversarial loss $\mathcal{L}_{adv}$ are presented as bellow:
    \begin{equation}
    \label{eq:cls_and_adv}
    \begin{split}
        \mathcal{L}_{cls} 
        & =  
        \mathbb{E}_{(\mathbf{x}_s, \mathbf{y}_s) \sim \mathcal{D}_{\mathcal{S}}} 
        \mathcal{L}_{ce} ( C (  F( \mathbf{x}_s ) ), \mathbf{y}_s)
        \\
        \mathcal{L}_{adv} 
        & =  
        \mathbb{E}_{\mathbf{f}_s \sim \tilde{\mathcal{D}}_s} 
        \log \left[D\left( F (\mathbf{x}_s) \right)\right] \\
        & + 
        \mathbb{E}_{\mathbf{f}_t \sim \tilde{\mathcal{D}}_t} 
        \log \left[1 - D\left( F (\mathbf{x}_t) \right)\right]
    \end{split}
    \end{equation}
    where $ \tilde{\mathcal{D}}_s $ and $\tilde{\mathcal{D}}_t $ denote the induced feature distributions of $\mathcal{D}_s$ and $\mathcal{D}_t$ respectively, 
    and $\mathcal{L}_{ce} (\cdot , \cdot)$ is the cross-entropy loss function.

\subsection{Label Propagation}
    To effectively utilize the unlabeled data, pseudo-labeling methods have been employed in both semi-supervised learning
    \cite{iscen2019label, lee2013pseudo, liu2019learning, wu2018unsupervised} and 
    domain adaptation scenarios \cite{liang2021domain, zhang2020progressive, wang2021uncertainty}.
    These methods pick up the class with the maximum predicted probability as true labels each time the weights are updated.
    Based on similarity graphs, it is intuitive to adopt the Label Propagation (LPA) to generate pseudo-labels with the homophily assumption widely used in graph neural networks that nearby examples get the same predictions \cite{mcpherson2001birds}.
    Directly applying LPA usually depends on solving the linear system 
    $\mathbf{Z} = (\mathbf{I} - \pi \mathbf{D}^{-1/2} \mathbf{A} \mathbf{D}^{-1/2})^{-1} \mathbf{Y}$, 
    where
    $\pi \in \left[ 0, 1\right)$ is a parameter,
    $\mathbf{A} \in \mathbb{R}^{n \times n} $ is the adjacency matrix of the graph constructed by $n$ data samples,
    $\mathbf{D}$ is the corresponding degree matrix,
    $\mathbf{I}$ is the identity matrix, 
    and $\mathbf{Y} \in \mathbb{R}^{n \times c}$ is the label matrix corresponding to all labeled examples.
    Finally, this method can obtain the pseudo-label $\hat{\mathbf{y}}_i = \arg \max_j \mathbf{z}_{i j}$, 
    where $\mathbf{Z}_{i j}$ is the element of matrix $\mathbf{Z}$.

\section{Methods}
    
    Our approach enables us to effectively capture the underlying domain distributions while ensuring that the learned features are transferable across domains.
    The overall pipeline is shown in Figure \ref{fig:framework}.
    Specifically, based on our constructed dynamic graphs in Section \ref{sec:graph_construction}, we propose a novel framework that
    utilizes graph spectra to align inter-domain relations
    in Section \ref{sec:spectral},
    and leverages intra-domain relations via a neighbor-aware propagation mechanism in Section \ref{sec:smooth}.
    Finally, we give the summary of each module and final objectives of \ourloss{} and \ours{} in Section \ref{sec:approach}.
    \ours{} can adapt to UDA, SSDA, MSDA, MTDA, and even solve specific data adaptation challenges.

\subsection{Dynamic Graph Construction}  \label{sec:graph_construction}
    Images and text data inherently contain rich sequential or spatial structures that can be effectively represented using graphs. 
    By constructing graphs based on the data, 
    both intra-domain relations and inter-domain relations can be exploited. 
    For instance,
    semantic and spatial relations among detected objects in an image \cite{li2019relation}, 
    cross-modal relations among images and sentences \cite{cheng2022cross}, intra-domain relations among images and texts \cite{chen2020graph, iscen2019label} have been successfully modeled using graph-based representations.    
    In this paper, leveraging self-correlation graphs enables us to 
    model the relations between different samples within a domain and capture the underlying data distributions.
    
    Our dynamic graphs are constructed on source features and target features, respectively.
    Firstly, we yield these extracted feature representations $\mathbf{f}_s$ from the source domain, and we aim to construct an undirected and weighted source graph 
    $\mathcal{G}_s = (\mathcal{V}_s, \mathcal{E}_s)$.
    Each vertex $v_i \in \mathcal{V}_s$ is represented by a feature representation $\mathbf{f}_i^s$.
    Each weighted edge $e_{i j} \in \mathcal{E}_s$ can be formulated as a relation between a pair of entities $\phi( \mathbf{f}_i^s, \mathbf{f}_j^s )$, 
    where $\phi(\cdot)$ denotes a type of metric function.
    Similarly, we also obtain extracted feature representations $\mathbf{f}_t$ from the target domain, and then, the target graph $\mathcal{G}_t = (\mathcal{V}_t, \mathcal{E}_t)$ can be constructed in the same way.

    Note that both $\mathbf{f}_s$ and $\mathbf{f}_t$ keep evolving 
    along with the update of model parameters during the training process.
    The metric function $\xi(\cdot)$ used in our implementation is cosine similarity, Gaussian similarity, and Euclidean distance.    
    The node feature matrix corresponding to the source and target graphs is denoted as $\mathbf{X}_s$ and $\mathbf{X}_t$, 
    and adjacency matrix are denoted as $\mathbf{A}_s$ and $\mathbf{A}_t$, respectively.
    As is well-known, the node feature matrix of a graph contains node-local information, and the adjacency matrix contains all of its topological information.
    
    After our dynamic graph construction, the problem shifts to addressing how to learn transferable features within the graph-based DA framework. 
    This involves designing methodologies that effectively leverage the graph structure to facilitate feature transfer. 
    To establish a generalization bound on our defined dynamic graph, it is essential to first introduce the concept of the $k$-hop ego graph. 
    Let's proceed to define the k-hop ego graph formally.
        
    \begin{definition}
    \label{def:khop}
    \textsc{ ($k$-hop Ego-graph \cite{zhu2021transfer}) }.
        Let $\mathcal{G}_i = (\mathcal{V}_i, \mathcal{E}_i) $ be a $k$-hop ego-graph centered at node $v_i$ if it has a $k$-layer centroid expansion \cite{bai2016fast} s.t. the greatest distance between $v_i$ 
        any other nodes in the ego-graph is $k$, i.e. $\forall v_j \in \mathcal{V}_i,\left| d\left(v_i, v_j\right) \right| \leq k$, where $d\left(v_i, v_j\right)$ is the graph distance between $v_i$ and $v_j$.
    \end{definition}

    \begin{theorem}
    \label{the:Bound}
    \textsc{ (Generalization bound for Graph-based DA) }.
        Suppose that the learned discriminator $g$ is $C_g$-Lipschitz continuous and the graph feature extractor $f$ is $C_f$-Lipschitz. 
        Let $\mathcal{H}:=\{h: \mathcal{G} \rightarrow \mathcal{Y}\}$ be the set of bounded real-valued functions with the pseudo-dimension as Pdim($\mathcal{H}$) = $d$ that $h = g \circ f \in \mathcal{H}$ with probability at least $1 - \delta$.
        The following inequality holds:
        \begin{equation*}
        \begin{aligned}
            \epsilon_t (h) 
            & 
            \leq 
            \hat{\epsilon}_s (h)
            + \sqrt{
                \frac{4 d}{N_s} \log (\frac{e N_s}{d} )
                + \frac{1}{N_s} \log (\frac{1}{\delta} )
                } \\
            & + 2 C_g C_f w( \mathbb{P} (\mathcal{G}_s), \mathbb{P} (\mathcal{G}_t) ) 
            + \eta 
        \end{aligned}
        \end{equation*}
        where the empirical source risk is 
        $
        \hat{\epsilon}_s (h) 
        = \frac{1}{N_s} \| h(\mathcal{G}_s) - \hat{h}(\mathcal{G}_s) \|
        $ 
        and the target risk is
        $
        \epsilon_t (h) 
        = \mathbb{E}_{\mathbb{P}(\mathcal{G}_t)} 
        (\| h(\mathcal{G}_t) - \hat{h}(\mathcal{G}_t) \|)
        $ 
        with the true labeling function for graphs $\hat{h}: \mathcal{G} \rightarrow \mathcal{Y}$.
        Here $w(\cdot, \cdot)$ is the Wasserstein distance
        and the probability distribution of a graph $\mathbb{P} (\mathcal{G})$
        is defined as the distribution of all the ego-graphs of $\mathcal{G}$.
        And
        $
        \eta = \min _{h \in \mathcal{H}} 
        \left\{ 
        \epsilon_s  (h^*) + \epsilon_t (h^*)
        \right\}
        $
        denotes the optimal combined error that can be achieved on both source and target graphs by the optimal hypothesis.
    \end{theorem}
    
Theorem \ref{the:Bound} provides a domain adaptation generalization bound based on graph-data structures. 
As observed, the first two terms are related to the performance on the source domain. 
It is a common practice to enhance source domain performance to achieve better outcomes on the target domain. 
The second term increases at first, and then decreases as the size of the source graph further grows.
Our dynamic graph is limited by the batch size, resulting in a small graph. 
This is consistent with the finding that a smaller source graph can achieve similar generalization performance when it helps reduce the distribution divergence.

The smaller the distribution divergence between the source and target graphs, \emph{i.e.}, $w(\mathbb{P}\left(\mathcal{G}_s\right), \mathbb{P}\left(\mathcal{G}_t\right))$, the smaller the generalization bound.
Note that Wasserstein distance can be extended to other metric distances.
According to \cite{redko2020survey, sejdinovic2013equivalence}, the Euclidean distance and other metrics can be adapted through the  Reproducing Kernel Hilbert Space (RKHS) framework, allowing this theorem to apply more broadly.

\subsection{Graph Spectral Alignment}  \label{sec:spectral}
    From the previous theory, a key aspect of graph-based DA is reducing the discrepancy between the source and target domains. Inter-domain alignment is often framed as a graph matching problem. However, explicit graph matching methods involve complex combinatorial optimization to find one-to-one correspondences between nodes or edges \cite{chen2020graph}. Our goal, instead, is to align the distributions of the source and target domains without such intricate processes.
    
    To achieve this, we utilize an implicit alignment method in the spectral space. By leveraging the properties of graph Laplacians, we align the source and target graphs in the spectral space, avoiding the need for explicit graph matching. This approach allows for a smoother and more efficient domain adaptation process, as the source and target features are aligned into the same eigenspace during learning \cite{zhuo2023towards}, thereby reducing domain discrepancy without complex matching stages. 
    Next, we provide the definition of the graph Laplacians.
    
\begin{definition}
\label{def:graph_laplacians}
\textsc{(Graph Laplacians \cite{luxburg2007tutorial})}.
    Let $\mathcal{G} = (\mathcal{V}, \mathcal{E}) $ be a finite graph with vertices $\mathcal{V}$ and weighted edges $\mathcal{E}$.
    Let $\phi: \mathcal{V} \rightarrow \mathcal{R} $ be a function of the vertices taking values in a ring
    and $\gamma : \mathcal{E} \rightarrow \mathcal{R} $ be a weighting function of weighed edges.
    Then, the graph Laplacian $\Delta$ acting on $\phi$ and $\gamma$ is defined by 
    $$(\Delta_{\gamma} \phi)(v)=\sum_{w: d(w, v)=1} \gamma_{wv}[\phi(v)-\phi(w)]$$
    where $d(w, v)$ is the graph distance between vertices $w$ and $v$, 
    and $\gamma_{wv}$ is the weight value on the edge $wv \in \mathcal{E}$.
\end{definition}
    Definition \ref{def:graph_laplacians} is a crucial operator in graph signal processing.
    Graph Laplacian filters apply a smoothing operation to the signal on a graph by taking advantage of the local neighborhood structure of the graph represented by the Laplacian matrix \cite{luxburg2007tutorial, brouwer2011spectra}, allowing effective propagation of node information.
    Building on this concept, we will provide a theoretical analysis to further explore the transferability between source and target graphs.

\begin{theorem}
\label{the:Transferability}
\textsc{(Transferability)}.
    Given a source graph $\mathcal{G}_s$ and a target graph $\mathcal{G}_t$, the transferability of graph feature extractor $f$ satisfies 
    $$    
    \| f(\mathcal{G}_s) - f(\mathcal{G}_t) \| 
    \leq 
    C_1 \Delta (\mathcal{G}_s, \mathcal{G}_t) + C_2
    $$
    where  $C_1$ and $C_2$ are two positive constants, 
    and 
    $
    \Delta (\mathcal{G}_s, \mathcal{G}_t)
    = \frac{1}{N_s N_t} 
    \sum_{i=1}^{N_s} \sum_{j=1}^{N_t} 
    \| \mathbf{L}_{\mathcal{G}^s_i} - \mathbf{L}_{\mathcal{G}^t_j} \|
    $ 
    measures the graph Laplacian distance between $\mathcal{G}_s$ and $\mathcal{G}_t$.
    Here $\mathbf{L}_{\mathcal{G}^s_i}$ and $\mathbf{L}_{\mathcal{G}^t_j}$ indicate the normalized graph Laplacian based on the ego-graph of node $i$ and node $j$ in source graph $\mathcal{G}_s$ and target graph $\mathcal{G}_t$, respectively.
\end{theorem}
    Theorem \ref{the:Transferability} establishes a transferability bound for a graph feature extractor between the source and target graphs. 
    The bound suggests that transferable features from the source to the target domain are directly influenced by the differences between the normalized graph Laplacians based on ego-graphs of nodes in the source and target graphs.
    According to spectral graph theory \cite{chung1997spectral}, there are important properties of graph Laplacians and their eigenvalues, \emph{i.e.}, algebraic connectivity and spectral gap \cite{wang2023message}.
    Here, a smaller matrix norm indicates a closer alignment between the source and target graphs, leading to better transferability.
    Then, we delve into the matrix norm of the graph Laplacians to provide further insights into the spectral alignment between source and target graphs.

\begin{definition}
\label{def:laplacian_spectra}
\textsc{(Spectral Distances)} .
    Given two simple and nonisomorphic graphs
    $\mathcal{G}_s$ and $\mathcal{G}_t$
    on $n$ vertices with the spectra of Laplacians 
    $\Lambda_s = \left\{\lambda_i^s\right\}_{i=1}^n$ with 
    $\lambda_1^s \ge \lambda_2^s \ge \cdots \ge \lambda_n^s $
    and
    $\Lambda_t = \left\{\lambda_i^t\right\}_{i=1}^n$ with 
    $\lambda_1^t \ge \lambda_2^t \ge \cdots \ge \lambda_n^t $
    respectively.
    Define the spectral distance between $\mathcal{G}_s$ and $\mathcal{G}_t$ as
    $$ d_{\Lambda} (\mathcal{G}_s, \mathcal{G}_t) = \Vert \Lambda_s - \Lambda_t \Vert_p , \quad p \ge 1 $$
\end{definition}
    Building on the previous theoretical insights, we define a spectral distance to measure the discrepancy between the source and target graphs in the spectral space. 
    The distance between the spectra of graphs can effectively quantify how far the spectrum of one graph with $n$ vertices is from that of another graph with the same number of vertices, which is often explored in the mathematical literature \cite{aouchiche2014distance, stevanovic2007research, gu2015spectral} and can also be related to the $p\,$th-Wasserstein distance, providing a simplified means to evaluate the discrepancy of graphs. 

    For a simple undirected graph, Definition \ref{def:graph_laplacians} corresponds directly to the Laplacian matrix.
    Given the adjacency matrix $\mathbf{A}_s$ of source graph $\mathcal{G}_s$,
    we derive its Laplacian matrix $\mathbf{L}_s$ and corresponding eigenvalues $\Lambda_s$. 
    Similarly, for target graph $\mathcal{G}_t$, we obtain its Laplacian matrix $\mathbf{L}_t$ and eigenvalues $\Lambda_t$.
    Following Definition \ref{def:laplacian_spectra}, 
    the spectral distance between $\mathcal{G}_s$ and  $\mathcal{G}_t$ is calculated.
    We define the graph spectral alignment penalty $\alpha$ as:
    \begin{equation}
    \label{eq:spectral_loss}
    \begin{split}
    \mathcal{L}_{gsa} & = 
    d_{\Lambda} (\mathcal{G}_s, \mathcal{G}_t)
    \end{split}
    \end{equation}
    This spectral penalty measures the discrepancy between the source and target graphs in the spectral space.
    Minimizing this penalty reduces the spectral distance between two domains, facilitating better domain alignment.
    It can also serve as a regularization term that is easily integrated with existing domain adaptation methods and can be effectively extended to various domain adaptation tasks and scenarios.

\begin{theorem}
\label{the:subspace}
\textsc{(Spectral Subspace Override Bound)}.
Given two matrices 
$ \mathbf{L}_t, \mathbf{L}_s \in \mathbb{R}^n $ with $n > 1$ 
and the rank of 
$ \mathbf{L}_t $ and $ \mathbf{L}_s > 1 $,
the norm 
$ \| \mathbf{L}_s^l - \mathbf{L}_t^l \| \in \mathbb{R}^l $ 
induced by the normalized subspace projector
$ \mathbf{M} \in \mathbb{R}^{n \times l} $ with $ \mathbf{M}^T \mathbf{M} = \mathbf{I} $ is bounded by
$$ 
\| \mathbf{L}_s^l - \mathbf{L}_t^l \|
< 
\sum_{i=1}^{l+1} 
(\Lambda_s^i - \Lambda_t^i )^2 
\leq 
\| \mathbf{L}_s - \mathbf{L}_t \|
$$
\end{theorem}
    Insight by the subspace override bound in \cite{raab2020low}, we provide Theorem \ref{the:subspace}, which refines the matrix norm differences between the graph Laplacians by focusing on their eigenvalues. 
    It connects spectral properties to our graph spectral alignment, and further, we define spectral distance.

\subsection{Smoothing Consistency Analysis}  \label{sec:smooth}
    In this section, we will introduce how to exploit intra-domain relations within target domain graphs. The well-trained source domain naturally forms tight clusters in the latent space. Since the spectra of the two domains are aligned, the rich topological information is coarsely transferred to the target domain. To perform fine-grained intra-domain alignment, we take a further step in encouraging message propagation within the target graph. To formalize this process, we present a new theorem.
    
\begin{theorem}
\label{the:lpa}
\textsc{(Smoothing Consistency)}.
Suppose that the latent ground-truth mapping: 
$\mathcal{M} : \mathcal{X} \rightarrow \mathcal{Y}$ from node features to node labels is differentiable and satisfies $C_m$-Lipschitz constant,
\emph{i.e}, 
$ 
|\mathcal{M}(\mathbf{x}_1) - \mathcal{M}(\mathbf{x}_2) | 
\leq 
C_m \| \mathbf{x}_1 - \mathbf{x}_2 \| 
$ 
for any $\mathbf{x}_1$ and $\mathbf{x}_2$. 
If edge weight $A_{i j}$ approximately smooth $\mathbf{x}_i$ over its immediate neighbors with error $\epsilon_i$, \emph{i.e},
$
\mathbf{x}_i = 
    \frac{1}{\mathbf{D}_{i i}} 
    \sum_{j \in \mathcal{N}_i} 
    \mathbf{A}_{i j} \mathbf{x}_j r
    + \epsilon_i 
$,
then the edge weight $A_{i j}$ also approximately smooth $y_i$ 
over its immediate neighbors with the following approximation error:
$$
|\mathbf{y}_i - 
 \frac{1}{\mathbf{D}_{ii}} 
 \sum_{j \in \mathcal{N}_i} 
\mathbf{A}_{i j} \mathbf{y}_j |
\leq 
C_m \| \epsilon_i \| + 
o( \textstyle \max_{ j \in \mathcal{N}_i } \| \mathbf{x}_j - \mathbf{x}_i \| )
$$
where $o(\alpha)$ denotes a higher order infinitesimal than $\alpha$ .
\end{theorem}

    Theorem \ref{the:lpa} addresses the concept of smoothing consistency through Label Propagation (LPA) following \cite{wang2020unifying}. 
    The aforementioned Theorem \ref{the:Bound} and Theorem \ref{the:Transferability} are based on a graph feature extractor which usually uses classic graph neural networks \cite{bruna2014spectral, defferrard2016convolutional, kipf2017semi}. 
    However, such networks can be overly complex and computationally intensive for image data. Additionally, given that our dynamic graphs are small in size, we prefer a more lightweight approach. 
    A non-parametric method, LPA, offers a more efficient alternative for message propagation within the graph. 
    Theorem \ref{the:lpa} just ensures that using LPA for label smoothing inherently leads to feature smoothing. 
    It shows that if the edge weights $\mathcal{A}_{i j}$ approximately smooth a node feature $\mathbf{x}_i$ over its immediate neighbors, they also approximately smooth the node label $y_i$ with a bounded approximation error. 
    
    This result establishes LPA as a suitable method for achieving feature consistency, providing a simpler yet effective means of propagating information within the target graph.
    Note that solving the linear system of LPA is performed on all data samples, 
    and implicitly requires access to all embedding vectors at each step of computation \cite{wu2018unsupervised}.
    If the size of the dataset increases, 
    it would be intractable to finish the matrix computation.
    Based on our dynamic graphs mentioned in Section \ref{sec:graph_construction},
    we only focus on unlabeled target graphs for each iteration and thus adapt the classic LPA to an iterative version with the assistance of a memory bank.

    \vpara{Memory Bank.}
    The memory bank stores each target data sample's prediction probabilities along with its associated feature vector, using the target sample's index as the key.
    To reduce ambiguity in these target predictions, we refine the model's output predictions by applying a sharpening technique combined with class balancing \cite{berthelot2019mixmatch, guo2017calibration}.
    In particular, we raise each predicted probability to the power of $-\tau$ and then re-normalize:
    $$
    \begin{aligned}
       \tilde{\mathbf{p}}_{i, c}^m 
        & =  
        \mathbf{p}_{i, c}^{-\tau}  / \textstyle 
        \sum_{i} \mathbf{p}_{i, c}^{-\tau} 
    \end{aligned}
    $$
    where $\mathbf{p}_i = C ( F ( \mathbf{x}_i ) )$ is the predicted probability vector for sample $\mathbf{x}_i$ so that $\hat{\mathbf{y}}_i = \arg \max_{k} \mathbf{p}_{i, k}$, and $\tau$ is a temperature scaling factor.
    As $\tau \rightarrow 0$, this sharpening drives the distribution to collapse to a point mass \cite{lee2013pseudo}.
    Additionally, we normalize the sharpened predictions using the overall class distribution vector to ensure balanced class representation across the unlabeled target data.
    Finally, the memory bank is updated at each iteration via an exponential moving average (EMA) with decay rate $\xi$.
    
    \vpara{Neighbor-aware Propagation.}
    For each sample $\mathbf{x}_i$ in the current mini-batch, we retrieve its $k$ nearest neighbors from the memory bank based on cosine similarity between the sample's feature $F(\mathbf{x}_i)$ and all stored features $\mathbf{f}_j^m$ in the memory. 
    We then aggregate the soft predictions of these neighbors by averaging them. 
    Formally, the neighbor-averaged prediction $\mathbf{q}_{i,c}$ for class $c$ is computed as:
    $$
    \mathbf{q}_{i,c} = \frac{1}{k} 
    \textstyle 
    \sum_{j \neq i, j \in \mathcal{N}_i} \tilde{\mathbf{p}}^m_{j, c}
    $$
    where $\mathcal{N}_i$ is the index set of the $k$ nearest neighbors of $x_i$ in the memory. Different neighborhoods $\mathbf{N}_i$ may have varying local densities, so samples residing in a denser region will naturally produce higher $\mathbf{q}_{i,c}$ values for their predicted class \cite{liang2021domain}.
    In other words, a larger $\mathbf{q}_{i,c}$ indicates that $\mathbf{x}_i$ lies in a high-density neighborhood, implying higher confidence in its pseudo-label. 
    Thus, we treat $\mathbf{q}_{i,c}$ as a confidence score for the pseudo-label of $\mathbf{x}_i$. 
    Accordingly, we define a confidence-weighted cross-entropy loss over all $N_t$ target samples as:
    \begin{equation}
    \label{eq:pl_loss}
    \begin{split}
        \mathcal{L}_{nap} 
        & = 
        - \frac{1}{N_t} \textstyle 
        \sum_{i=1}^{N_t} \mathbf{q}_{i, \hat{\mathbf{y}}_i} 
        \log \mathbf{p}_{i, \hat{\mathbf{y}}_i}
    \end{split}
    \end{equation}

\subsection{Overall Architecture} \label{sec:approach}

    \vpara{\ourloss{}.} 
    The \ourloss{} consists of both graph spectral alignment and neighbor-aware propagation after dynamic graph construction.
    Given $\mathcal{L}_{gsa}$ in Eq.\ref{eq:spectral_loss} and neighbor-aware propagation  $\mathcal{L}_{nap}$ in Eq.\ref{eq:pl_loss}, 
    the final objective of \ourloss{} can be yielded as:
    \begin{equation}
    \label{eq:spa_loss}
        \mathcal{L}_{SPA\_Loss} =  
        \alpha \cdot \mathcal{L}_{gsa} + 
        \beta \cdot \mathcal{L}_{nap}
    \end{equation}
    where $\alpha$ is the coefficient term to control the regularization effects of the graph spectral alignment mechanism, 
    and $\beta$ is the coefficient term properly designed to grow along with iterations
    to mitigate the noises in the pseudo-labels at early iterations and avoid the error accumulation \cite{lee2013pseudo}.
    Concerning the labeled data, we implement the standard cross-entropy loss with label-smoothing regularization \cite{szegedy2016rethinking}.
    We also give different trials of similarity metrics and graph Laplacians.

    \vpara{\ours{}.} 
    To solve more adaptation scenarios, we further extend \ourloss{} via data augmentation and progressive consistency.
    For unlabeled target samples in each mini-batch and $\mathbf{p}_i = C ( F ( \mathbf{x}_i ) )$ represents an image $\mathbf{x}_i$ in the mini-batch.
    Also, $\mathbf{p}_i' = C ( F ( \mathbf{x}_i' ) )$ indicates the prediction of a transformed image  $\mathbf{x}_i'$, which is an augmented version of $\mathbf{x}_i'$ using a data augmentation technique. And then, the inner product $\mathbf{p}_i\transpose{} \mathbf{p}_i'$ is used as a similarity score, which predicts whether image $\mathbf{x}_i$ and the transformed version of image $\mathbf{x}_j$ share the same class label or not. 
    Thus, the $\mathbf{p}_i$ in Eq. \ref{eq:pl_loss} can be replaced with the enhanced $\mathbf{p}_i'$ with a consistency control.
    Furthermore, with the augmented features of target samples, we can construct another augmented graph $\mathcal{G}_a$, making alignment between the source graph and this augmented graph.
    These data augmentation techniques combined in the process of pairwise comparison can significantly strengthen the model performance \cite{li2021cross}.
    In this way, the final objective of \ours{} can be formulated as:
    \begin{equation}
        \mathcal{L}_{SPA++} =  
        \alpha \cdot \mathcal{L}_{gsa++} + 
        \beta \cdot \mathcal{L}_{nap++} +
        \mathcal{L}_{con}
    \end{equation}
    where 
    \begin{equation*}
        \mathcal{L}_{con} = \omega(t) \sum_{i=1}^{N_t} \| \mathbf{p}_i - \mathbf{p}_i'  \|_2     
    \end{equation*}
    and $\omega(t) = ve^{-5(1-\frac{t}{T})^2}$ is a ramp-up function used in \cite{laine2017temporal} with the scaler coefficient $v$, the current step $t$ and the total steps $T$ in the ramp-up process. 
    To further adapt \ourloss{} to more scenarios, we augment target samples to generalize the graph spectral alignment and neighbor-aware propagation loss.
    Therefore, we employ the $\mathcal{L}_{con}$ to keep the consistency of the augmented samples.
    The details of generalized graph spectral alignment and neighbor-aware propagation with the auto-augmented target graph $\mathcal{G}_a$ and prediction $\mathbf{p}_i'$ of augmented samples $\mathbf{x}_i'$ are as follows:
    \begin{equation*}
        \mathcal{L}_{gsa++} = d_{\Lambda} (\mathcal{G}_s, \mathcal{G}_t) + d_{\Lambda} (\mathcal{G}_s, \mathcal{G}_a)
    \end{equation*}
    \begin{equation*}
        \mathcal{L}_{nap++}  = - \sum_{i=1}^{N_t} 
        \mathbf{1} \{ \text{max}(\mathbf{p}_i) \ge c \}   
        \cdot
        \mathbf{q}_{i, \hat{\mathbf{y}}_i}  
        \log \mathbf{p}_{i, \hat{\mathbf{y}}_i}'
    \end{equation*}
    Note that we only describe the problem of unsupervised domain adaptation in this section for simplicity, to give a detailed theoretical proof.
    In practice, our \ours{} can easily extend to more adaptation scenarios. 
    More concrete details are in the following section.

\section{Experiments} \label{sec:experiment}
This section presents a comprehensive evaluation of our proposed method, structured as follows.
Section \ref{sec:setup} details the experimental configurations, including benchmark datasets, data processing workflows, and implementation specifics such as hyperparameters and toolbox.
Section \ref{sec:exp_uda} reports the model performance under inductive UDA and default transductive UDA standards, demonstrating our effectiveness.
Section \ref{sec:exp_vda} offers the results of more adaptation scenarios such as SSDA, MSDA, MTDA, and specific distribution challenges, showing our generalization and robustness of adaptability.
Section \ref{sec:exp_model} gives a comprehensive model analysis including ablation studies, parameter sensitivity, diverse visualizations, etc.

\subsection{Experimental Setups} \label{sec:setup}
\begin{table}[b]
\centering
\caption{The detailed description of datasets. Note that DomainNet126 and DomainNet are different versions of the same data resource.}
\label{tab:dataset_detail}
\resizebox{0.48\textwidth}{!}{
    \begin{tabular}{l|lll}
        \toprule
        Dataset & Domains & \#Classes & \#Images \\
        \midrule
        Office31 & A, W, D & 31 & 4110 \\
        OfficeHome & A, C, P, R & 65 & 15500 \\
        VisDA2017 & Synthetic, Real & 12 & 280157 \\
        DomainNet126 & C, P, R, S & 126 & 145145 \\
        DomainNet & C, I, P, Q, R, S & 345 & 569010 \\
        \bottomrule
    \end{tabular}
}
\end{table}
\vpara{Datasets.} 
We conduct experiments on a series of benchmark datasets, as summarized in Table \ref{tab:dataset_detail}:

\begin{itemize}
    \item{
        \textbf{Office31} \cite{saenko2010adapting} is a widely-used benchmark for visual DA. It contains 4,652 images of 31 office environment categories from three domains: \textit{Amazon} (A), \textit{DSLR} (D), and \textit{Webcam} (W), which correspond to online websites, digital SLR cameras, and web camera images.
    }
    \item{
        \textbf{OfficeHome} \cite{venkateswara2017deep} is a challenging dataset that consists of images of everyday objects from four different domains: 
        \textit{Artistic} (A), \textit{Clipart} (C), \textit{Product} (P), and \textit{Real-World} (R). 
        Each domain contains 65 object categories in office and home environments, amounting to 15,500 images around. 
        Following the typical setting in \cite{chen2019transferability}, 
        we evaluate methods on one-source to one-target domain adaptation scenarios, 
        resulting in 12 adaptation cases in total.
    }
    \item{
        \textbf{VisDA2017} \cite{peng2018visda} is a large-scale benckmark that attempts to bridge the significant synthetic-to-real domain gap with over 280,000 images across 12 categories. 
        The source domain has 152,397 synthetic images generated by rendering from 3D models. 
        The target domain has 55,388 real object images collected from \textit{Microsoft COCO} \cite{lin2014microsoft}.
        Following \cite{na2021fixbi}, 
        we evaluate methods on synthetic-to-real tasks and present test accuracy for each category.
    }
    \item{
        \textbf{DomainNet} \cite{peng2019moment} is a large-scale dataset containing about 600,000 images across 345 categories, which span 6 domains with a large domain gap: 
        \textit{Clipart} (C), \textit{Infograph} (I), \textit{Painting} (P), \textit{Quickdraw} (Q), \textit{Real} (R), and \textit{Sketch} (S).
        Following the UDA settings in \cite{tllib}, we focus on 12 tasks among C, P, R, and S domains on the original DomainNet dataset for an inductive UDA setting. 
    }
    \item{ 
        \textbf{DomainNet126} \cite{peng2019moment} is a simplified version of the aforementioned DomainNet.
        As labels of some domains and classes are very noisy, this version picked samples from 4 domains (\textit{Real}, \textit{Clipart}, \textit{Painting}, \textit{Sketch}) and 126 classes. It focuses on the adaptation scenarios where the target domain is not real images and constructs 7 scenarios from the four domains.
        Following the UDA settings in \cite{tllib}, we also use DomainNet126 for the default transductive UDA setting.
        Following the SSDA settings in \cite{saito2019semi}, we focus on 1-shot and 3-shot on the simplified DomainNet126 dataset.
    }
\end{itemize}

    \vpara{Implementation details.} 
    We use PyTorch and tllib toolbox \cite{tllib} to implement our method and fine-tune ResNet pre-trained on ImageNet \cite{he2016deep, he2016identity}. 
    Following the standard protocols for unsupervised domain adaptation in previous methods \cite{liang2021domain, na2021fixbi}, we use the same backbone networks for fair comparisons. 
    For Office31 and OfficeHome datasets, we use ResNet-50 as the backbone network. 
    For the VisDA2017 and DomainNet datasets, we use ResNet-101 as the backbone network.
    Following previous work \cite{liang2021domain}, we adopt mini-batch stochastic gradient descent to learn the feature encoder by fine-tuning from the ImageNet pre-trained model with the learning rate of 0.001, and new layers, as a bottleneck and classification layer.
    The learning rates of the layers trained from scratch are set to be 0.01. 
    We use the same learning rate schedule in \cite{liang2021domain, long2018conditional}, including a learning rate scheduler with a momentum of 0.9, a weight decay of 0.005, a bottleneck size of 256, and a batch size of 32. 
    We report the main experimental results with the average accuracy over 5 random trials.
    For transductive unsupervised domain adaptation, the reported accuracy is computed on the complete unlabeled target data, following the established protocol for UDA \cite{jin2020minimum, chen2019transferability, liang2021domain}.
    For inductive unsupervised domain adaptation on DomainNet,
    the reported accuracy is computed on the provided test dataset.
    We use a standard batch size of 32 for both source and target in all experiments and for all variants of our method.
    The reverse validation \cite{liang2019exploring, zhong2010cross} is conducted to select hyperparameters.
    For both UDA and SSDA scenarios, we fix the coefficient of $\mathcal{L}_{nap}$ as 0.2 and the coefficient of $\mathcal{L}_{gsa}$ as 1.0, while we will offer a sensitivity analysis for these two coefficients in the following section.
    For multi-domain and specific data distribution scenarios, we fix the coefficient of $\mathcal{L}_{nap}$ as 0.2 and the coefficient of $\mathcal{L}_{gsa}$ as 0.5.

\begin{table*}[!b]
\centering
\caption{
Classification Accuracy (\%) on the DomainNet dataset for the inductive UDA setting, 
using the ResNet101 model as the backbone. 
Note that in this inductive setting, we compare methods on the original DomainNet dataset with train and test splits in the target domain, leading to an inductive scenario.
The \textbf{bold} text indicates the highest accuracy in each column,
and the \textcolor{blue}{blue} text indicates the accuracy improvement of our method compared to the second-highest one among others.
}
\label{tab:uda_domainnet_rn101}
\resizebox{0.98\textwidth}{!}{
    \begin{tabular}{l|cccccccccccc|l}
    \toprule 
    Method & 
    C$\rightarrow$P & C$\rightarrow$R & C$\rightarrow$S & 
    P$\rightarrow$C & P$\rightarrow$R & P$\rightarrow$S & 
    R$\rightarrow$C & R$\rightarrow$P & R$\rightarrow$S & 
    S$\rightarrow$C & S$\rightarrow$P & S$\rightarrow$R & Avg. \\
    \midrule 
    \textit{Source} \cite{he2016deep} &
    32.7 & 50.6 & 39.4 & 41.1 & 56.8 & 35.0 & 48.6 & 48.8 & 36.1 & 49.0 & 34.8 & 46.1 
    & 43.3 \\
    DANN \cite{ganin2015unsupervised} &
    37.9 & 54.3 & 44.4 & 41.7 & 55.6 & 36.8 & 50.7 & 50.8 & 40.1 & 55.0 & 45.0 & 54.5 
    & 47.2 \\
    CDAN \cite{long2018conditional} &
    39.9 & 55.6 & 45.9 & 44.8 & 57.4 & 40.7 & 56.3 & 52.5 & 44.2 & 55.1 & 43.1 & 53.2 
    & 49.1 \\
    MCD \cite{saito2018maximum} &
    37.5 & 52.9 & 44.0 & 44.6 & 54.5 & 41.6 & 52.0 & 51.5 & 39.7 & 55.5 & 44.6 & 52.0 
    & 47.5 \\
    MCC \cite{jin2020minimum} & 
    40.1 & 56.5 & 44.9 & 46.9 & 57.7 & 41.4 & 56.0 & 53.7 & 40.6 & 58.2 & 45.1 & 55.9 
    & 49.7 \\
    MDD \cite{li2020maximum} &
    42.9 & 59.5 & 47.5 & 48.6 & 59.4 & 42.6 & 58.3 & 53.7 & 46.2 & 58.7 & 46.5 & 57.7 
    & 51.8 \\
    SDAT \cite{rangwani2022closer} &
    41.5 & 57.5 & 47.2 & 47.5 & 58.0 & 41.8 & 56.7 & 53.6 & 43.9 & 58.7 & 48.1 & 57.1 
    & 51.0 \\
    Leco \cite{wang2022revisiting} &
    44.1 & 55.3 & 48.5 & 49.4 & 57.5 & 45.5 & 58.8 & 55.4 & 46.8 & 61.3 & 51.1 & 57.7 
    & 52.6 \\
    \midrule
    \rowcolor{Gray} 
    \ourloss{} &
    54.3          & 70.9 & 56.1 & 
    \textbf{59.3} & 71.5 & 51.8 & 
    64.6          & 59.6 & 52.1 & 
    66.0          & 57.4 & 70.6 & 
    61.2 \textcolor{blue}{(+8.6)}
    \\ 
    \rowcolor{Gray}
    \ours{} & 
    \textbf{55.3} & \textbf{71.6} & \textbf{56.6} & 
    58.3          & \textbf{72.2} & \textbf{53.1} & 
    \textbf{66.1} & \textbf{61.3} & \textbf{54.6} & 
    \textbf{66.8} & \textbf{58.2} & \textbf{71.1} & 
    \textbf{62.1} \textcolor{blue}{(+9.5)} \\
    \bottomrule
    \end{tabular}
}
\end{table*}

\begin{table*}[!b]
\centering
\caption{
Classification Accuracy (\%) on DomainNet126 dataset for default transductive UDA setting, using the ResNet101 model as the backbone. 
The \textbf{bold} text indicates the highest accuracy in each column,
and the \textcolor{blue}{blue} text indicates the accuracy improvement of our method compared to the second-highest one among others.
}
\label{tab:uda_multi_rn101}
    \resizebox{0.98\textwidth}{!}{
        \begin{tabular}{l|cccccccccccc|l}
        \toprule 
        Method & 
        C$\rightarrow$P & C$\rightarrow$R & C$\rightarrow$S & 
        P$\rightarrow$C & P$\rightarrow$R & P$\rightarrow$S & 
        R$\rightarrow$C & R$\rightarrow$P & R$\rightarrow$S & 
        S$\rightarrow$C & S$\rightarrow$P & S$\rightarrow$R & Avg. \\
        \midrule 
        \textit{Source} \cite{he2016deep} & 
        38.4 & 50.9 & 43.9 & 50.3 & 66.7 & 39.9 & 54.6 & 57.9 & 43.7 & 52.5 & 43.5 & 48.3 &
        49.2 \\
        DANN \cite{ganin2015unsupervised} & 
        46.5 & 58.2 & 51.6 & 52.7 & 64.2 & 52.9 & 61.7 & 60.3 & 53.9 & 62.7 & 56.7 & 61.6 &
        56.9 \\
        CDAN \cite{long2018conditional} &
        50.9 & 61.6 & 54.8 & 59.4 & 68.5 & 55.5 & 70.4 & 66.9 & 57.7 & 64.2 & 59.1 & 64.3 &
        61.1 \\
        MCD \cite{saito2018maximum} & 
        43.7 & 55.7 & 47.6 & 51.9 & 67.8 & 45.0 & 52.9 & 57.3 & 40.4 & 56.3 & 50.8 & 56.8 & 
        52.3 \\
        BSP \cite{chen2019transferability} & 
        45.7 & 58.7 & 55.5 & 48.6 & 65.2 & 48.6 & 55.2 & 60.8 & 48.6 & 56.8 & 55.8 & 61.4 &
        55.1 \\
        SAFN \cite{xu2019larger} & 
        50.0 & 58.7 & 52.4 & 56.3 & 73.7 & 53.5 & 55.8 & 64.8 & 48.5 & 60.7 & 59.5 & 64.3 & 
        58.2 \\
        RSDA \cite{gu2020spherical} & 
        45.5 & 56.6 & 46.6 & 45.7 & 60.4 & 48.6 & 54.6 & 61.5 & 50.9 & 56.1 & 54.0 & 58.6 & 
        53.4 \\
        PAN \cite{wang2020progressive} & 
        58.8 & 65.2 & 54.6 & 57.5 & 70.5 & 53.1 & 67.6 & 66.7 & 55.9 & 64.4 & 60.2 & 66.6 &
        61.8 \\
        MemSAC \cite{kalluri2022memsac} &
        53.6 & 66.5 & 58.8 & 63.2 & 71.2 & 58.1 & 73.2 & 70.5 & 61.5 & 68.8 & 64.1 & 67.6 & 
        64.8 \\
        \midrule
        \rowcolor{Gray} 
        \ourloss{} & 
        73.5 & 84.0 & 70.6 & 76.5 & 85.9 & 71.9 & \textbf{76.6} & 77.0 & 69.8 & \textbf{78.3} & 76.8 & 83.9 & 
        77.1 \textcolor{blue}{(+12.3)} 
        \\
        \rowcolor{Gray} 
        \ours{} & 
        \textbf{73.6} & \textbf{84.1} & \textbf{71.3} & \textbf{77.0} & \textbf{86.0} & \textbf{73.3} & 77.5 & \textbf{77.1} & \textbf{70.8} & 78.0 & \textbf{77.5} & \textbf{84.8} & \textbf{77.6} \textcolor{blue}{(+12.8)} 
        \\
        \bottomrule
        \end{tabular}
    }
\end{table*}

\begin{table*}[!t]
\centering
\caption{
Classification Accuracy (\%) on OfficeHome dataset for default transductive UDA setting, using ResNet50 as the backbone. 
The \textbf{bold} text indicates the highest accuracy in each column,
and the \textcolor{blue}{blue} text indicates the accuracy improvement of our method compared to the second-highest one among others.
}
\label{tab:uda_officehome_rn50}
\resizebox{0.98\textwidth}{!}{
    \begin{tabular}{l|cccccccccccc|l}
    \toprule 
    Method & 
    A$\rightarrow$C & A$\rightarrow$P & A$\rightarrow$R & 
    C$\rightarrow$A & C$\rightarrow$P & C$\rightarrow$R & 
    P$\rightarrow$A & P$\rightarrow$C & P$\rightarrow$R & 
    R$\rightarrow$A & R$\rightarrow$C & R$\rightarrow$P & Avg. \\
    \midrule 
    \textit{Source} \cite{he2016deep} &
    34.9 & 50.0 & 58.0 & 37.4 & 41.9 & 46.2 & 38.5 & 31.2 & 60.4 & 53.9 & 41.2 & 59.9 & 
    46.1 \\
    DANN \cite{ganin2015unsupervised} &
    45.6 & 59.3 & 70.1 & 47.0 & 58.5 & 60.9 & 46.1 & 43.7 & 68.5 & 63.2 & 51.8 & 76.8 &
    57.6 \\
    CDAN \cite{long2018conditional} & 
    50.7 & 70.6 & 76.0 & 57.6 & 70.0 & 70.0 & 57.4 & 50.9 & 77.3 & 70.9 & 56.7 & 81.6 &
    65.8 \\
    BSP \cite{chen2019transferability} &
    52.0 & 68.6 & 76.1 & 58.0 & 70.3 & 70.2 & 58.6 & 50.2 & 77.6 & 72.2 & 59.3 & 81.9 &
    66.3 \\
    NPL \cite{lee2013pseudo} & 
    54.1 & 74.1 & 78.4 & 63.3 & 72.8 & 74.0 & 61.7 & 51.0 & 78.9 & 71.9 & 56.6 & 81.9 & 
    68.2 \\ 
    MCC \cite{jin2020minimum} &
    56.3 & 77.3 & 80.3 & 67.0 & 77.1 & 77.0 & 66.2 & 55.1 & 81.2 & 73.5 & 57.4 & 84.1 & 
    71.0 \\
    BNM \cite{cui2020towards} &
    56.7 & 77.5 & 81.0 & 67.3 & 76.3 & 77.1 & 65.3 & 55.1 & 82.0 & 73.6 & 57.0 & 84.3 &
    71.1 \\
    NWD \cite{chen2022reusing} & 
    58.1 & 79.6 & 83.7 & 67.7 & 77.9 & 78.7 & 66.8 & 56.0 & 81.9 & 73.9 & 60.9 & 86.1 &  72.6 \\
    FixBi \cite{na2021fixbi} &
    58.1 & 77.3 & 80.4 & 67.7 & 79.5 & 78.1 & 65.8 & 57.9 & 81.7 & 76.4 & 62.9 & 86.7 &
    72.7 \\
    \midrule
    \rowcolor{Gray} 
    \ourloss{} &
    60.4 & 79.7 & 84.5 & 73.6 & 81.3 & 82.1 & 72.2 & 58.0 & 85.2 & 77.4 & 61.0 & 88.1 & 
    75.3 \textcolor{blue}{(+2.6)} 
    \\
    \rowcolor{Gray} 
    \ours{} & 
    \textbf{61.1} & \textbf{79.8} & \textbf{85.1} & \textbf{76.0} & \textbf{83.0} & \textbf{83.9} & \textbf{76.6} & \textbf{62.3} & \textbf{85.8} & \textbf{80.1} & \textbf{63.6} & \textbf{88.6} & \textbf{77.2} \textcolor{blue}{(+4.5)}  \\
    \bottomrule
    \end{tabular}
    }
\end{table*}

\begin{table*}[!t]
\centering
\caption{
Per-category Accuracy (\%) on the VisDA2017 dataset for the default transductive UDA setting, using ResNet101 as the backbone. 
The \textbf{bold} text indicates the highest accuracy in each column,
and the \textcolor{blue}{blue} text indicates the accuracy improvement of our method compared to the second-highest one among others.
}
\label{tab:uda_visda_rn101}
\resizebox{0.98\textwidth}{!}{
    \begin{tabular}{l|cccccccccccc|l}
    \toprule
    Method &
    aero & bike & bus & car &
    horse & knife & motor & person &
    plant & skate & train & truck & Avg. \\
    \midrule
    \textit{Source} \cite{he2016deep} & 
    55.1 & 53.3 & 61.9 & 59.1 & 80.6 & 17.9 & 79.7 & 31.2 & 81.0 & 26.5 & 73.5 &  8.5 & 
    52.4 \\
    DANN \cite{ganin2015unsupervised}      & 
    81.9 & 77.7 & 82.8 & 44.3 & 81.2 & 29.5 & 65.1 & 28.6 & 51.9 & 54.6 & 82.8 &  7.8 &  
    57.4 \\
    CDAN \cite{long2018conditional}        &
    85.2 & 66.9 & 83.0 & 50.8 & 84.2 & 74.9 & 88.1 & 74.5 & 83.4 & 76.0 & 81.9 & 38.0 & 
    73.7 \\
    BSP \cite{chen2019transferability}     & 
    92.4 & 61.0 & 81.0 & 57.5 & 89.0 & 80.6 & 90.1 & 77.0 & 84.2 & 77.9 & 82.1 & 38.4 & 
    75.9 \\
    NPL \cite{lee2013pseudo}               & 
    90.9 & 74.6 & 73.2 & 55.8 & 89.6 & 64.6 & 86.8 & 68.7 & 90.7 & 64.8 & 89.5 & 47.7 &
    74.7 \\
    MCC \cite{jin2020minimum}              & 
    92.2 & 82.9 & 76.8 & 66.6 & 90.9 & 78.5 & 87.9 & 73.8 & 90.1 & 76.1 & 87.1 & 41.0 &
    78.8 \\
    BNM \cite{cui2020towards}              & 
    93.6 & 68.3 & 78.9 & 70.3 & 91.1 & 82.8 & 93.0 & 78.7 & 90.9 & 76.5 & 89.1 & 40.9 & 
    79.5 \\
    NWD \cite{chen2022reusing}             & 
    96.1 & 82.7 & 76.8 & 71.4 & 92.5 & 96.8 & 88.2 & 81.3 & 92.2 & 88.7 & 84.1 & 53.7 & 
    83.7 \\
    FixBi \cite{na2021fixbi}               &
    96.1 & 87.8 & 90.5 & 90.3 & 96.8 & 95.3 & 92.8 & 88.7 & 97.2 & 94.2 & 90.9 & 25.7 & 
    87.2 \\
    \midrule
    \rowcolor{Gray} 
    \ourloss{} & 
    98.5 & 92.2 & 86.3 & 63.0 & 97.5 & 95.4 & 93.5 & 80.7 & 97.2 & 95.2 & 91.1 & 61.4 &
    87.7 \textcolor{blue}{(+0.5)} 
    \\ 
    \rowcolor{Gray} 
    \ours{} & 
    \textbf{98.6} & \textbf{92.5} & \textbf{87.0} & \textbf{63.8} & \textbf{98.2} & \textbf{95.9} & \textbf{94.5} & \textbf{81.6} & \textbf{97.9} & \textbf{96.1} & \textbf{92.0} & \textbf{63.2} & \textbf{88.4} \textcolor{blue}{(+1.2)}
    \\
    \bottomrule
    \end{tabular}
    }
\end{table*}

\begin{table}[!ht]
\centering
\caption{
Classification Accuracy (\%) on Office31 dataset for default transductive UDA setting, using ResNet50 as the backbone.
The \textbf{bold} text indicates the highest accuracy in each column,
and the \textcolor{blue}{blue} text indicates the accuracy improvement of our method compared to the second-highest one among others.
}
\label{tab:uda_office31_rn50}
     \resizebox{0.49\textwidth}{!}{
        \begin{tabular}{l|cccccc|l}
        \toprule 
        Method & 
        A$\rightarrow$D & A$\rightarrow$W &
        D$\rightarrow$A & D$\rightarrow$W & 
        W$\rightarrow$A & W$\rightarrow$D & Avg. \\
        \midrule 
        \textit{Source} \cite{he2016deep} &
        78.3 & 70.4 & 57.3 & 93.4 & 61.5 & 98.1 & 76.5 \\
        DANN \cite{ganin2015unsupervised} &
        79.7 & 82.0 & 68.2 & 96.9 & 67.4 & 99.1 & 82.2 \\
        CDAN \cite{long2018conditional} & 
        92.9 & 94.1 & 71.0 & 98.6 & 69.3 & 100. & 87.7 \\
        BSP \cite{chen2019transferability} & 
        93.0 & 93.3 & 73.6 & 98.2 & 72.6 & 100. & 88.5 \\ 
        NPL \cite{lee2013pseudo} & 
        88.7 & 89.1 & 65.8 & 98.1 & 66.6 & 99.6 & 84.7 \\
        MCC \cite{jin2020minimum} &
        92.1 & 94.0 & 74.9 & 98.5 & 75.3 & 100. & 89.1 \\
        BNM \cite{cui2020towards} &
        92.2 & 94.0 & 74.9 & 98.5 & 75.3 & 100. & 89.2 \\
        NWD \cite{chen2022reusing} & 
        95.4 & 95.2 & 76.4 & 99.1 & 76.5 & 100. & 90.4 \\
        FixBi \cite{na2021fixbi} &
        95.0 & 96.1 & 78.7 & 99.3 & 79.4 & 100. & 91.4 \\
        \midrule
        \rowcolor{Gray} 
        \ourloss{} &
        95.0 & 97.2 & 78.0 & \textbf{99.0} & 79.4 & 99.8 & 91.4 \\ 
        \rowcolor{Gray}        
        \ours{} & 
        \textbf{96.8} & \textbf{97.4} & \textbf{79.4} & 98.9 & \textbf{79.9} & \textbf{100.} & \textbf{92.1} \textcolor{blue}{(+0.7)}
        \\
        \bottomrule
        \end{tabular}
    }
\end{table}

\subsection{Unsupervised Domain Adaptation} \label{sec:exp_uda}

In this section, we compare \ours{} with various state-of-the-art methods for unsupervised domain adaptation (UDA) scenarios, including inductive and transductive settings, where \textit{Source} in the UDA task means the model trained only using labeled source data. 

    \vpait{Inductive UDA.}
    Table \ref{tab:uda_domainnet_rn101} shows the results on the DomainNet dataset.
    We compare \ourloss{} and \ours{} with the various state-of-the-art methods for inductive UDA scenarios on this large-scale dataset.
    The inductive setting follows that we directly use the train and test splits of target domain data as the original dataset \cite{peng2019moment}. 
    During the training process, we can use the train split of the target domain data, and then we use the test split of the target domain data for testing.
    This setting is relatively more difficult than others, with the large-scale data amount and inductive learning.
    Only a few works follow this setting, while we give a result comparison in this setting here. 
    The experiments show \ourloss{} and \ours{} consistently outperform various DA methods with an average accuracy of 61.2\% and 62.1\%, which are 8.6\% and 9.5\% higher than the recent second-best work, Leco \cite{wang2022revisiting}, respectively. 
    This demonstrates the superiority of \ourloss{} and \ours{} in this inductive setting.
    
    \vpait{Transductive UDA.} 
    Transductive UDA is the default UDA setting followed by most of the previous works.
    Table \ref{tab:uda_multi_rn101}, Table \ref{tab:uda_officehome_rn50}, Table \ref{tab:uda_visda_rn101} and Table \ref{tab:uda_office31_rn50}
    show the results on DomaiNet126, OfficeHome, VisDA2017 and Office31 datasets respectively.
    We compare \ourloss{} and \ours{} with the various state-of-the-art methods for the default transductive UDA scenarios on these public benchmarks.
    In the aforementioned section, we have presented the classification accuracy results on the original DomainNet with 365 categories.
    While the original DomainNet dataset has noisy labels, the previous work \cite{saito2019semi} uses a subset of it that contains 126 categories from C, P, R, and S, 4 domains in total, which we refer to as DomainNet126. 
    Thus, here, we report the results of the transductive UDA setting on the DomainNet126 dataset.
    Table \ref{tab:uda_multi_rn101} (DomainNet126) shows the average accuracy of \ourloss{} and \ours{} is 77.1\% and 77.6\%.
    Further, compared with classic baselines, 
    we improve DANN \cite{ganin2015unsupervised} by 20.7\%, and CDAN \cite{long2018conditional} by 16.8\%.
    Note that we directly follow the tuned hyperparameters based on the OfficeHome dataset and did not do further cumbersome parameter searches on the DomainNet126 dataset, showing the generality of our method to some extent.
    Table \ref{tab:uda_officehome_rn50} (OfficeHome) shows the average accuracy of \ourloss{} and \ours{} is 75.3\% and 77.2\%, achieving the best accuracy,
    which is 2.6\% and 4.5\% higher than the previous method FixBi \cite{na2021fixbi}. 
    Note that the average accuracy reported by NWD is based on MCC \cite{jin2020minimum} while \ourloss{} and \ours{} are based on DANN \cite{ganin2015unsupervised}.
    DANN is a more classic and simple baseline, showing our great improvement.
    Table \ref{tab:uda_visda_rn101} (VisDA2017) shows the average classification accuracy of \ourloss{} and \ours{} is 87.7\% and 88.4\%, achieving the best accuracy, which is 0.5\% and 1.2\% higher than the previous method FixBi \cite{na2021fixbi}.
    Looking at this table, we report the details of per-category accuracy, and we can find that our methods consistently outperform most of the domain adaptation methods. 
    Specifically, for \ourloss{} on the VisDA2017 dataset, we adopt a basic mixup for randomly cropping input images to achieve an accuracy of 87.7\%.
    While for \ours{} on the VisDA2017 dataset, we directly follow our pipeline without any special input processing, since the \ours{} includes a data augmentation and makes augmented copies consistent, leading to better accuracy with the simple pipeline.
    Table \ref{tab:uda_office31_rn50} (Office31) is on the small-sized Office31 dataset.
    The average accuracy of \ourloss{} and \ours{} is 91.4\% and 92.1\%, respectively, outperforming most DA methods.
    Note that \ourloss{} achieves comparable results with FixBi \cite{na2021fixbi} and \ours{} achieves the highest performance, showing the efficacy of our auto-augmentation and progressive consistency strategies.


    \begin{table*}[!t]
    \centering
    \caption{
    Classification Accuracy (\%) on the DomainNet126 dataset for 1-shot and 3-shot SSDA setting, 
    using ResNet34 as the backbone. 
    The \textbf{bold} text indicates the highest accuracy in each column,
    and the \textcolor{blue}{blue} text indicates the accuracy improvement of our method compared to the second-highest one among others.
    }
    \label{tab:ssda_domainnet}
        \resizebox{0.98\textwidth}{!}{
        \begin{tabular}{l|cccccccl|cccccccl}
        \toprule 
        \multirow{2}{*}{Method} & \multicolumn{8}{c|}{1-shot} & \multicolumn{8}{c}{3-shot} \\
         & 
         C$\rightarrow$S & P$\rightarrow$C & P$\rightarrow$R & R$\rightarrow$C & R$\rightarrow$P & R$\rightarrow$S & S$\rightarrow$P & Avg. & 
         C$\rightarrow$S & P$\rightarrow$C & P$\rightarrow$R & R$\rightarrow$C & R$\rightarrow$P & R$\rightarrow$S & S$\rightarrow$P & Avg. \\
        \midrule 
        \textit{S + T} \cite{he2016deep} & 
        54.8 & 59.2 & 73.7 & 61.2 & 64.5 & 52.0 & 60.4 & 60.8 & 
        57.9 & 63.0 & 75.6 & 63.9 & 66.3 & 56.0 & 62.2 & 63.6 \\
        DANN \cite{ganin2015unsupervised} & 
        52.8 & 70.3 & 56.3 & 58.2 & 61.4 & 52.2 & 57.4 & 58.4 & 
        55.4 & 72.2 & 59.6 & 59.8 & 62.8 & 54.9 & 59.9 & 60.7 \\
        ENT \cite{grandvalet2004semi} &
        54.6 & 65.4 & 75.0 & 65.2 & 65.9 & 52.1 & 59.7 & 62.6 & 
        60.0 & 71.1 & 78.6 & 71.0 & 69.2 & 61.1 & 62.1 & 67.6 \\
        MME \cite{saito2019semi} & 
        56.3 & 69.0 & 76.1 & 70.0 & 67.7 & 61.0 & 64.8 & 66.4 & 
        61.8 & 71.7 & 78.5 & 72.2 & 69.7 & 61.9 & 66.8 & 68.9 \\
        MixMatch \cite{berthelot2019mixmatch} & 
        59.3 & 66.7 & 74.8 & 69.4 & 67.8 & 62.5 & 66.3 & 66.7 & 
        62.7 & 68.7 & 78.8 & 72.6 & 68.8 & 65.6 & 67.1 & 69.2 \\
        MCC \cite{jin2020minimum} & 
        56.8 & 62.8 & 75.3 & 65.5 & 66.9 & 57.6 & 63.4 & 64.0 & 
        60.5 & 66.5 & 76.5 & 67.2 & 68.1 & 59.8 & 65.0 & 66.2 \\
        NPL \cite{lee2013pseudo} & 
        62.5 & 67.6 & 78.3 & 70.9 & 69.2 & 62.0 & 67.0 & 68.2 & 
        64.5 & 70.7 & 79.3 & 72.9 & 70.7 & 64.8 & 68.6 & 70.2 \\
        APE \cite{kim2020attract} & 
        56.7 & 72.9 & 76.6 & 70.4 & 70.8 & 63.0 & 64.5 & 67.6 & 
        63.1 & 76.7 & 79.4 & 76.6 & 72.1 & 67.8 & 66.1 & 71.7 \\
        \midrule
        \rowcolor{Gray} \ourloss{} &
        65.9 & 74.8 & 81.1 & 75.3 & 71.8 & \textbf{65.8} & \textbf{69.8} & 
        72.1 \textcolor{blue}{(+3.9)} & 
        67.0 & \textbf{76.5} & \textbf{82.3} & \textbf{76.0} & 72.2 & 67.2 & \textbf{71.1} & 
        73.2 \textcolor{blue}{(+1.5)} 
        \\
        \rowcolor{Gray} \ours{} &
        \textbf{66.2} & \textbf{74.9} & \textbf{81.2} & \textbf{75.5} & \textbf{71.9} & \textbf{65.8} & \textbf{69.8} & 
        \textbf{72.2} \textcolor{blue}{(+4.0)} & 
        \textbf{67.1} & \textbf{76.5} & \textbf{82.3} & \textbf{76.0} & \textbf{73.3} & \textbf{67.3} & \textbf{71.1} & 
        \textbf{73.4} \textcolor{blue}{(+1.7)}
        \\
        \bottomrule
        \end{tabular}
        }
    \end{table*}

    \begin{table*}[!t]
    \centering
    \caption{
    Classification Accuracy (\%) on OfficeHome dataset for 1-shot and 3-shot SSDA setting, using ResNet34 as the backbone. 
    The \textbf{bold} text indicates the highest accuracy in each column,
    and the \textcolor{blue}{blue} text indicates the accuracy improvement of our method compared to the second-highest one among others.
    }
    \label{tab:ssda_officehome}
        \resizebox{0.98\textwidth}{!}{
        \begin{tabular}{l|cccccccccccc|l}
        \toprule 
        Method & 
        A$\rightarrow$C & A$\rightarrow$P & A$\rightarrow$R & 
        C$\rightarrow$A & C$\rightarrow$P & C$\rightarrow$R & 
        P$\rightarrow$A & P$\rightarrow$C & P$\rightarrow$R & 
        R$\rightarrow$A & R$\rightarrow$C & R$\rightarrow$P & Avg. \\
        \midrule 
        \multicolumn{14}{c}{1-shot} \\
        \midrule 
        \textit{S + T}  \cite{he2016deep} &     
        52.1 & 78.6 & 66.2 & 74.4 & 48.3  & 57.2 & 69.8 & 50.9 & 73.8 & 70.0 & 56.3 & 68.1 & 
        63.8 \\
        DANN \cite{ganin2015unsupervised}  & 
        53.1 & 74.8 & 64.5 & 68.4 & 51.9 & 55.7 & 67.9 & 52.3 & 73.9 & 69.2 & 54.1 & 66.8 & 
        62.7 \\
        ENT \cite{grandvalet2004semi} & 
        53.6 & 81.9 & 70.4 & 79.9 & 51.9 & 63.0 & 75.0 & 52.9 & 76.7 & 73.2 & 63.2 & 73.6 &
        67.9 \\ 
        MME \cite{saito2019semi} & 
        61.9 & 82.8 & 71.2 & 79.2 & 57.4 & 64.7 & 75.5 & 59.6 & 77.8 & 74.8 & 65.7 & 74.5 &
        70.4 \\
        APE \cite{kim2020attract} & 
        60.7 & 81.6 & 72.5 & 78.6 & 58.3 & 63.6 & 76.1 & 53.9 & 75.2 & 72.3 & 63.6 & 69.8 & 
        68.9 \\
        CDAC \cite{li2021cross} & 
        61.9 & 83.1 & 72.7 & 80.0 & 59.3 & 64.6 & 75.9 & 61.2 & 78.5 & 75.3 & 64.5 & 75.1 &
        71.0 \\
        \midrule
        \rowcolor{Gray} \ourloss{} &
        \textbf{62.3} & 76.7 & 79.0 & \textbf{66.6} & \textbf{77.3} & \textbf{76.4} & 65.7 & \textbf{59.1} & 80.7 & 71.4 & \textbf{65.2}  & 84.1 & 72.0 \textcolor{blue}{(+1.0)} \\
        \rowcolor{Gray} \ours{} & 
        \textbf{62.3} & \textbf{77.4} & \textbf{79.1} & \textbf{66.6} & \textbf{77.3} & \textbf{76.4} & \textbf{66.1} & 59.0 & \textbf{80.8} & \textbf{74.1} & 62.8 & \textbf{85.0} & \textbf{72.2} \textcolor{blue}{(+1.2)} \\
        \midrule
        \multicolumn{14}{c}{3-shot} \\
        \midrule
        \textit{S + T}  \cite{he2016deep} & 
        55.7 & 80.8 & 67.8 & 73.1 & 53.8 & 63.5 & 73.1 & 54.0 & 74.2 & 68.3 & 57.6 & 72.3 & 
        66.2 \\
        DANN \cite{ganin2015unsupervised}  & 
        57.3 & 75.5 & 65.2 & 69.2 & 51.8 & 56.6 & 68.3 & 54.7 & 73.8 & 67.1 & 55.1 & 67.5 & 
        63.5 \\
        ENT \cite{grandvalet2004semi} &
        62.6 & 85.7 & 70.2 & 79.9 & 60.5 & 63.9 & 79.5 & 61.3 & 79.1 & 76.4 & 64.7 & 79.1 & 
        71.9 \\
        MME \cite{saito2019semi} & 
        64.6 & 85.5 & 71.3 & 80.1 & 64.6 & 65.5 & 79.0 & 63.6 & 79.7 & 76.6 & 67.2 & 79.3 & 
        73.1 \\
        APE \cite{kim2020attract} & 
        66.4 & 86.2 & 73.4 & 82.0 & 65.2 & 66.1 & 81.1 & 63.9 & 80.2 & 76.8 & 66.6 & 79.9 & 
        74.0 \\
        CDAC \cite{li2021cross} & 
        67.8 & 85.6 & 72.2 & 81.9 & 67.0 & 67.5 &  80.3  & 65.9 & 80.6 & 80.2 & 67.4 & 81.4 & 
        74.2 \\ 
        \midrule
        \rowcolor{Gray} \ourloss{} &
        63.1 & 81.0 & 80.2 & 68.5 & 81.7 & 77.5 & \textbf{69.5} & 65.2 & 82.0 & 73.9 & 67.2 & \textbf{87.0} & 
        74.7 \textcolor{blue}{(+0.5)} \\
        \rowcolor{Gray} \ours{} & 
        \textbf{65.7} & \textbf{81.4} & \textbf{80.6} & \textbf{68.6} & \textbf{81.9} & \textbf{78.5} & 69.0 & \textbf{65.5} & \textbf{82.1} & \textbf{74.8} & \textbf{68.6} & 86.3 & \textbf{75.3}
        \textcolor{blue}{(+1.1)} \\
        \bottomrule
        \end{tabular}
        }
    \end{table*}

\subsection{More Adaptation Scenarios} \label{sec:exp_vda}
    In this section, we verify that our method can be extended to versatile DA scenarios, including semi-supervised, multi-source, and multi-target  DA settings. 
    Furthermore, we also exploit special data distributions such as long-tail and subpopulation.

    \vpait{Semi-Supervised DA.} First, we extend our method to a semi-supervised domain adaptation (SSDA) scenario, and following \cite{saito2019semi}, we report the experimental results with 1-shot and 3-shot settings. 
    In the SSDA task, \textit{S + T} denotes the model trained only by the labeled source and target data.
    Especially, we present the classification accuracy results on DomainNet126 and OfficeHome datasets for the SSDA scenario in Table \ref{tab:ssda_domainnet} and Table \ref{tab:ssda_officehome}, respectively.
    Looking at the details,  
    Table \ref{tab:ssda_domainnet} shows the classification results for 1-shot and 3-shot SSDA settings on the DomainNet126 dataset.
    For the 1-shot setting, our SPA model can improve DANN \cite{ganin2015unsupervised} by 13.8\% and ENT \cite{grandvalet2004semi} by 9.6\%.
    For the 3-shot setting,
    our SPA model can improve DANN \cite{ganin2015unsupervised} by 12.7\% and ENT \cite{grandvalet2004semi} by 5.8\%.
    Furthermore, 
    Table \ref{tab:ssda_officehome} shows the classification results for 1-shot and 3-shot SSDA settings on the OfficeHome dataset.
    To verify that our SPA model can also generalize to the SSDA scenario, we compare SPA with several classic and recent baselines.
    The first section of the table shows our SPA model can improve DANN \cite{ganin2015unsupervised} by 9.5\% and ENT \cite{grandvalet2004semi} by 4.3\% in the 1-shot setting.
    The second section shows our SPA model can improve DANN \cite{ganin2015unsupervised} by 11.2\% and ENT \cite{grandvalet2004semi} by 2.8\% in the 3-shot setting.
    Overall, this part of the experimental results shows that our SPA model can greatly improve the classic baselines not only on the UDA setting but also on the SSDA setting.

    \begin{table}[!b]
    \centering
    \caption{
    Classification Accuracy (\%) on the DomainNet dataset for the MSDA setting, using ResNet101 as the backbone. 
    The \textbf{bold} text indicates the highest accuracy in each column,
    and the \textcolor{blue}{blue} text indicates the accuracy improvement of our method compared to the second-highest one among others.
    }
    \label{tab:msda_domainnet}
         \resizebox{0.49\textwidth}{!}{
            \begin{tabular}{l|cccccc|l}
            \toprule 
            Method(:t) & 
            $\rightarrow$C & $\rightarrow$I & $\rightarrow$P & 
            $\rightarrow$Q & $\rightarrow$R & $\rightarrow$S & Avg. \\
            \midrule 
            \textit{Source} \cite{he2016deep} &
            47.6 & 13.0 & 38.1 & 13.3 & 51.9 & 33.7 & 32.9 \\
            MCD  \cite{saito2018maximum}&
            54.3 & 22.1 & 45.7 & 7.6 & 58.4 & 43.5 & 38.5 \\
            DCTN  \cite{xu2018deep}&
            48.6 & 23.5 & 48.8 & 7.2 & 53.5 & 47.3 & 38.2 \\
            M$S^{3}$DA  \cite{peng2019moment}&
            58.6 & 26.0 & 52.3 & 6.3 & 62.7 & 49.5 & 42.6 \\
            CMSS \cite{yang2020curriculum} &
            64.2 & 28.0 & 53.6 & 16.0 & 63.4 & 53.8 & 46.5 \\
            LtC-MSDA  \cite{wang2020learning}&
            63.1 & 28.7 & 56.1 & 16.3 & 66.1 & 53.8 & 47.4 \\
            MCC  \cite{jin2020minimum}& 
            65.5 & 26.0 & 56.6 & 16.5 & 68.0 & 52.7 & 47.6 \\
            CDFA \cite{cai2023single} &
            68.3 & 23.1 & 55.2 & 12.4 & 69.4 & 55.2 & 47.3  \\
            CC-Loss  \cite{jin2024one}&
            67.4 & 28.6 & 58.6 & 18.0 & 72.4 & 56.0 & 50.2 \\
            \midrule
            \rowcolor{Gray} \ourloss{} &
            72.1 & 30.0 & 61.1 & 21.1 & 74.7 & 61.7 & 53.4 \textcolor{blue}{(+3.2)}
            \\ 
            \rowcolor{Gray} \ours{} &
            \textbf{74.6} & \textbf{31.6} & \textbf{63.8} & \textbf{27.2} & \textbf{77.2} & \textbf{62.9} & \textbf{56.2} \textcolor{blue}{(+6.0)}
            \\
            \bottomrule
            \end{tabular}
            }
    \end{table}
    \begin{table}[!b]
    \centering
    \caption{
    Classification Accuracy (\%) on the DomainNet dataset for the MTDA setting, using ResNet101 as the backbone. 
    The \textbf{bold} text indicates the highest accuracy in each column,
    and the \textcolor{blue}{blue} text indicates the accuracy improvement of our method compared to the second-highest one among others.
    }
    \label{tab:mtda_domainnet}
     \resizebox{0.49\textwidth}{!}{
            \begin{tabular}{l|cccccc|l}
            \toprule 
            Method(:t) & 
            C$\rightarrow$ & I$\rightarrow$ & P$\rightarrow$ & 
            Q$\rightarrow$ & R$\rightarrow$ & S$\rightarrow$ & Avg. \\
            \midrule 
            \textit{Source} \cite{he2016deep} &
            25.6 & 16.8 & 25.8 & 9.2 & 20.6 & 22.3 & 20.1 \\
            SE  \cite{french2017self}&
            21.3 & 8.5 & 14.5 & 13.8 & 16.0 & 19.7 & 15.6 \\
            MCD \cite{saito2018maximum} &
            25.1 & 19.1 & 27.0 & 10.4 & 20.2 & 22.5 & 20.7 \\
            DADA \cite{peng2019domain} &
            26.1 & 20.0 & 26.5 & 12.9 & 20.7 & 22.8 & 21.5 \\
            MCC  \cite{jin2020minimum}&
            33.6 & 30.0 & 32.4 & 13.5 & 28.0 & 35.3 & 28.8 \\
            CC-Loss  \cite{jin2024one}&
            36.1 & 33.3 & 34.8 & 22.2 & 27.7 & 37.4 & 31.9 \\
            \midrule
            \rowcolor{Gray} \ourloss{} &
            38.0 & \textbf{41.7} & 22.0 & \textbf{36.8} & 39.9 & 43.0 & 
            36.9 \textcolor{blue}{(+5.0)} \\ 
            \rowcolor{Gray} \ours{} & 
            \textbf{38.7} & 36.0 & \textbf{26.3} & 30.7 & \textbf{46.1} & \textbf{46.2} & 
            \textbf{37.3} \textcolor{blue}{(+5.4)} \\ 
            \bottomrule
            \end{tabular}
    }
    \end{table}
    \vpait{Multi-Source DA.} 
    When applying our method to the Multi-Source Domain Adaptation (MSDA) scenario, we merge multiple source domains and compare them with existing algorithms specially designed for MSDA on DomainNet. 
    In Table \ref{tab:msda_domainnet}, \textit{Source} in the MSDA scenario means the model trained only using labeled source data. 
    Following prior works \cite{peng2019moment}, we leave one domain out as the target domain and utilize the other domains as multiple source domains, leading to 6 adaptation tasks on the DomainNet dataset in total.
    Based on our spectral alignment mechanism, \ourloss{} and \ours{} outperform the existing strong methods specially designed for MSDA. 
    Note that these strong methods are designed with complex architecture and loss functions, which are relatively hard to use in practical applications, while ours can be plugged into other methods as a simple but effective regularization term with a significant improvement in model performance.

    \vpait{Multi-Target DA.} 
    We evaluate the Multi-Target Domain Adaptation (MTDA) tasks following the protocol of \cite{peng2019domain, jin2024one}, which provides six tasks on the DomainNet dataset. 
    We merge multiple target domains and compare them with existing algorithms specially designed for the MTDA setting on the DomainNet dataset. 
    As shown in Table \ref{tab:mtda_domainnet}, \textit{Source} in the MTDA scenario means the model trained only using labeled source data with the ResNet model. 
    Based on our spectral alignment mechanism, \ours{} outperforms the existing strong methods specially designed for MTDA. 
    Notably, the accuracy is very low on the source dataset, further highlighting the robustness of our method. 
    With well-designed mechanisms, it effectively mitigates the impact of incorrect predictions.

\begin{table}[!b]
\centering
\caption{
Classification Accuracy (\%) on OfficeHome (OH.) dataset for subpopulation shifts (Sub.) and long-tail setting.
The first section indicates single-domain and multi-domain long-tailed learning methods, and the second section includes approaches for improving robustness to distribution shift.
The \textbf{bold} text indicates the highest accuracy in each column,
and the \textcolor{blue}{blue} text indicates the accuracy improvement of our method compared to the second-highest one among others.
}
\label{tab:long_tail}
     \resizebox{0.49\textwidth}{!}{
        \begin{tabular}{l|l|ccccl}
        \toprule 
        \multirow{2}{*}{Method}
        & Sub. & \multicolumn{5}{c}{Multi-domain Long-tail} \\
        & OH. & A & C & P & R & Avg. \\
        \midrule 
        Focal \cite{ross2017focal}  & 62.6 &
        47.1 & 43.3 & 62.3 & 63.5 & 54.0  \\
        LDAM \cite{cao2019learning} & 63.6 &
        47.1 & 42.9 & 61.5 & 62.9 & 53.6  \\
        CRT \cite{kang2020decoupling} & 61.9 &
        47.2 & 42.6 & 61.4 & 63.3 & 53.6  \\
        MiSLAS \cite{zhong2021improving} & 61.4 &
        45.2 & 41.4 & 62.3 & 62.6 & 52.9  \\
        Remix \cite{chou2020remix} & 61.6 &
        44.3 & 39.2 & 60.7 & 61.6 & 51.4  \\
        RIDE \cite{raileanu2020ride} & 63.3 &
        47.2 & 43.0 & 61.5 & 63.4 & 53.8  \\
        PaCo \cite{cui2021parametric} & 63.0 &
        48.1 & 42.6 & 62.3 & 63.0 & 54.0   \\
        TALLY \cite{TALLY} & 
        \underline{67.0} &
        49.8 & 44.2 & 63.0 & 65.7 & \underline{55.7}  
        \\
        \midrule
        IRM \cite{arjovsky2019invariant} & 45.5 &
        33.6 & 34.3 & 49.5 & 52.0 & 42.3  \\
        GroupDRO \cite{sagawa2019distributionally} & 59.8 &
        44.6 & 41.8 & 58.4 & 59.6 & 51.1  \\
        CORAL \cite{sun2016deep}  & 59.1 &
        43.9 & 42.7 & 56.9 & 59.4 & 50.7  \\
        LISA \cite{yao2022improving} & 57.4  &
        41.8 & 37.0 & 56.5 & 57.6 & 48.2  \\
        MixStyle \cite{zhou2021mixstyle} & 62.3 &
        45.1 & 45.5 & 58.3 & 60.9 & 52.5  \\
        DDG \cite{zhou2021domain} & 58.8 &
        43.9 & 42.8 & 57.9 & 59.7 & 51.1  \\
        BODA \cite{yang2022multi} & 62.8 &
        47.1 & 44.3 & 59.6 & 62.3 & 53.3  \\
        \midrule
        \rowcolor{Gray}
        \ourloss{} & 
        69.5 \textcolor{blue}{(+2.5)} & 
        55.5 & 44.3 & 66.1 & 69.9 & 59.0 \textcolor{blue}{(+3.3)}
        \\
        \rowcolor{Gray}
        \ours{} & 
        \textbf{70.6} \textcolor{blue}{(+3.6)} & 
        \textbf{56.7} & \textbf{47.0} & 73.1 & 71.0 & 62.0 \textcolor{blue}{(+5.3)} 
        \\ 
        \bottomrule
        \end{tabular}
}
\end{table}

    \vpait{Long-tail.} 
    In the long-tail setting, we still follow the classical multi-domain setups. 
    We use one domain as the test domain, and the rest as the training domains. 
    The overall performance of \ours{} and prior methods for tackling multi-domain long-tail shifts is reported in Table \ref{tab:long_tail}.
    Single-domain long-tailed learning methods boost performance in most cases, showing that class imbalance is still an important issue in long-tail shifts.
    \ours{} achieves 62.0\% and \ourloss{} achieves 59.0\%, outperforming other methods with 5.3\% and 3.3\% error decreasing respectively.
    The results show that our methods consistently outperform prior approaches, indicating their efficacy in enhancing robustness to multi-domain long-tail shifts.

    \vpait{Subpopulation.} 
    In subpopulation shifts, the test set is balanced across domains and classes, which means that each domain-class pair contains the same number of text samples.
    The overall performance of \ours{} and prior methods for tackling subpopulation shift is reported in Table \ref{tab:long_tail}.
    For subpopulation shifts, we report the average performance over all domains.
    We first observe that most single-domain re-weighting approaches (e.g., Focal \cite{ross2017focal}, LDAM \cite{cao2019learning}) consistently outperform multi-domain learning approaches (e.g., GroupDRO \cite{sagawa2019distributionally}, CORAL \cite{sun2016deep}), indicating that imbalances in classes are probably more detrimental than imbalances in domains.
    This observation is not surprising, since all domains are observed during training and testing in subpopulation shift problems.
    Even so, \ours{} achieves 70.6\% and \ourloss{} achieves 69.5\%, outperforming other methods with 3.6\% and 2.5\% error decreasing respectively.
    This result verifies the effectiveness of our methods in improving the robustness of subpopulation shifts.
    Besides, it also indicates that designing regularizers that are suitable for datasets from diverse domains can be challenging. 
    Instead, balanced augmentation is capable of improving the robustness by transferring domain nuisances between examples.


\begin{table*}[!b]
\centering
\caption{
Ablation Study.
Classification accuracy (\%) on the OfficeHome dataset under the UDA setting.
We compare \ourloss{} and its variants with (w/) or without (w/o) $\mathcal{L}_{gsa}$, $\mathcal{L}_{nap}$, and also refer to the extended version \ours{}.
The first variant (regarded as baseline) omits both losses; \textcolor{blue}{blue} numbers show the accuracy gain of every other variant relative to this baseline.
These results show the effectiveness of both graph spectral alignment and neighbor-aware propagation mechanisms. 
Furthermore, the extended version via data augmentation and progressive consistency also achieves a good improvement.
}
\label{tab:ablation}
\resizebox{0.98\textwidth}{!}{
\begin{tabular}{l|cccccccccccc|l}
\toprule 
Method & 
A$\rightarrow$C & A$\rightarrow$P & A$\rightarrow$R & 
C$\rightarrow$A & C$\rightarrow$P & C$\rightarrow$R & 
P$\rightarrow$A & P$\rightarrow$C & P$\rightarrow$R & 
R$\rightarrow$A & R$\rightarrow$C & R$\rightarrow$P & Avg. \\
\midrule 
 \ourloss{} w/o $\mathcal{L}_{gsa}$, $\mathcal{L}_{nap}$ &
54.6 & 74.1 & 78.1 & 63.0 & 72.2 & 74.1 & 61.6 & 52.3 & 79.1 & 72.3 & 57.3 & 82.8 & 68.5 \\
 \ourloss{} w/o $\mathcal{L}_{gsa}$ &  
59.0 & 80.1 & 81.5 & 66.2 & 77.8 & 76.7 & 69.2 & 57.7 & 83.0 & 74.0 & 64.1 & 85.8 & 72.9 \textcolor{blue}{(+4.4)}\\
 \ourloss{} w/o $\mathcal{L}_{nap}$ & 
59.0 & 78.2 & 81.0 & 63.5 & 76.5 & 76.2 & 64.7 & 57.3 & 82.1 & 73.6 & 61.4 & 85.7 & 71.6 \textcolor{blue}{(+3.1)}\\
\ourloss{} & 
59.9 & 79.1 & 84.4 & 74.9 & 79.1 & 81.9 & 72.4 & 58.4 & 84.9 & 77.9 & 61.2 & 87.7 & 75.1 \textcolor{blue}{(+6.6)} \\ 
\ours{} & 
61.1 & 79.8 & 85.1 & 76.0 & 83.0 & 83.9 & 76.6 & 62.3 & 85.8 & 80.1 & 63.6 & 88.6 & 77.2 \textcolor{blue}{(+8.7)} \\
\bottomrule
\end{tabular}
}
\end{table*}

\begin{figure*}[!b]
\centering
\resizebox{\textwidth}{!}{
\begin{tabular}{ccc}
    \includegraphics[width=0.34\textwidth]{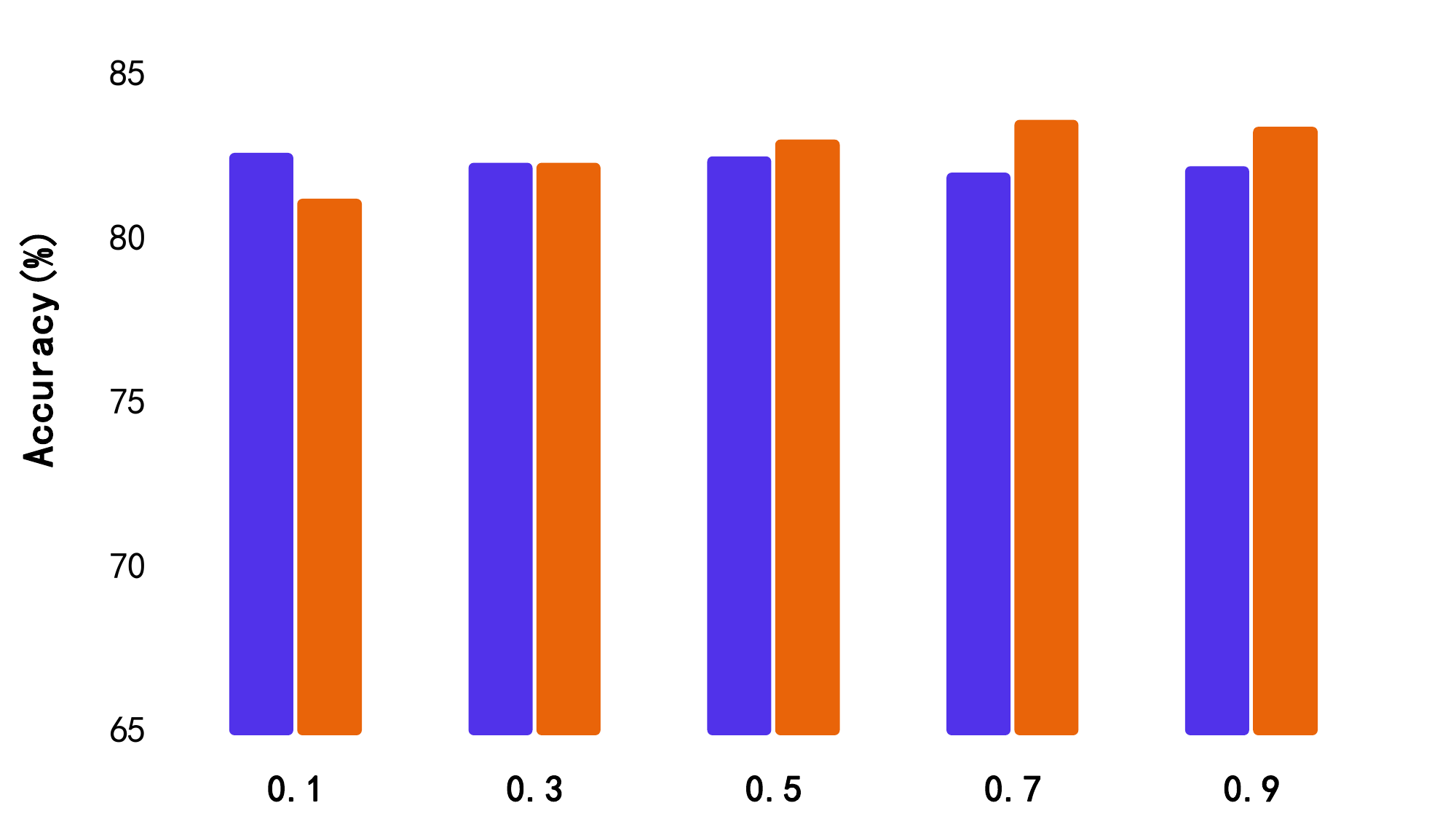} & 
    \includegraphics[width=0.30\textwidth]{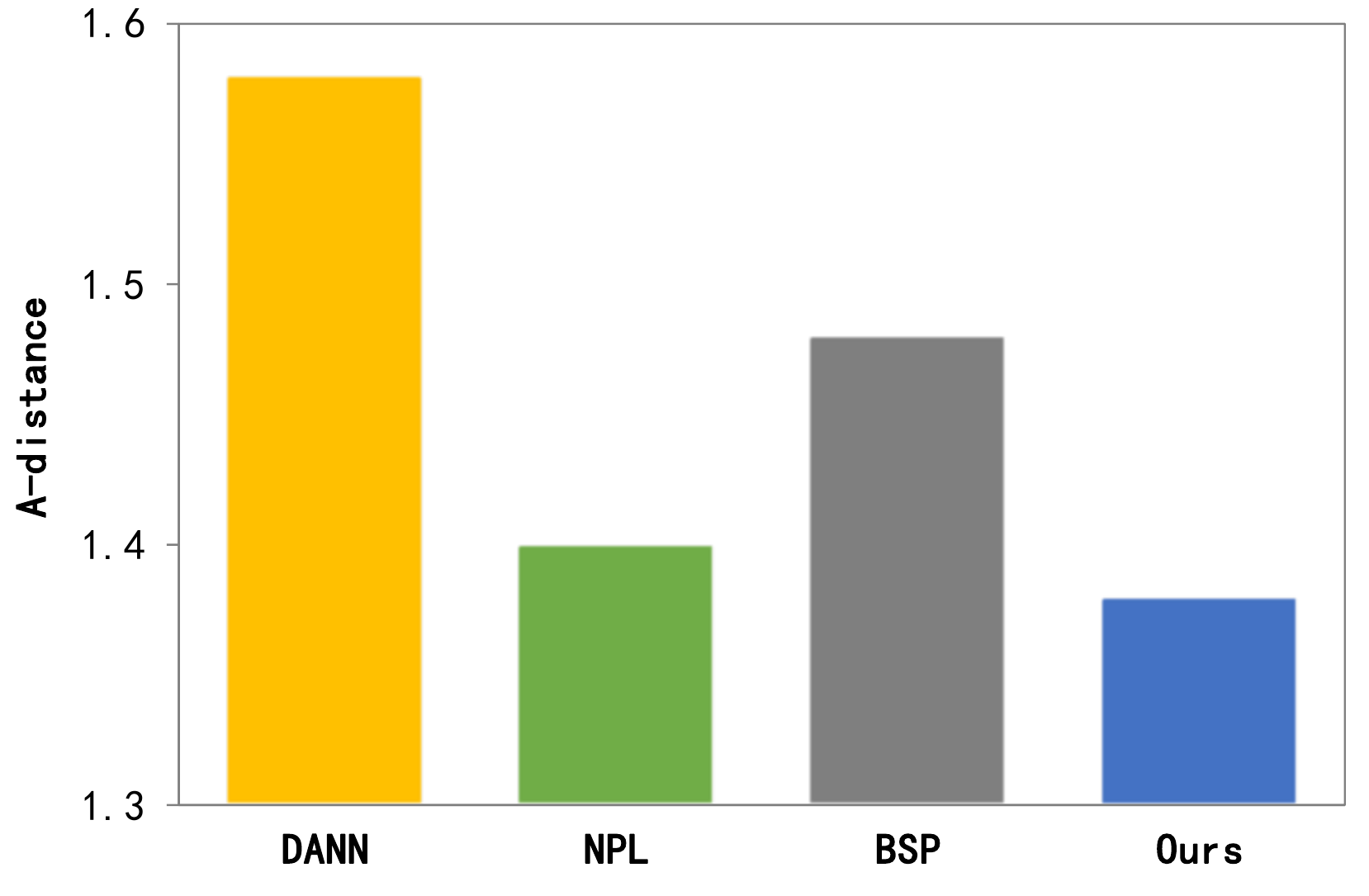} & 
    \includegraphics[width=0.32\textwidth]{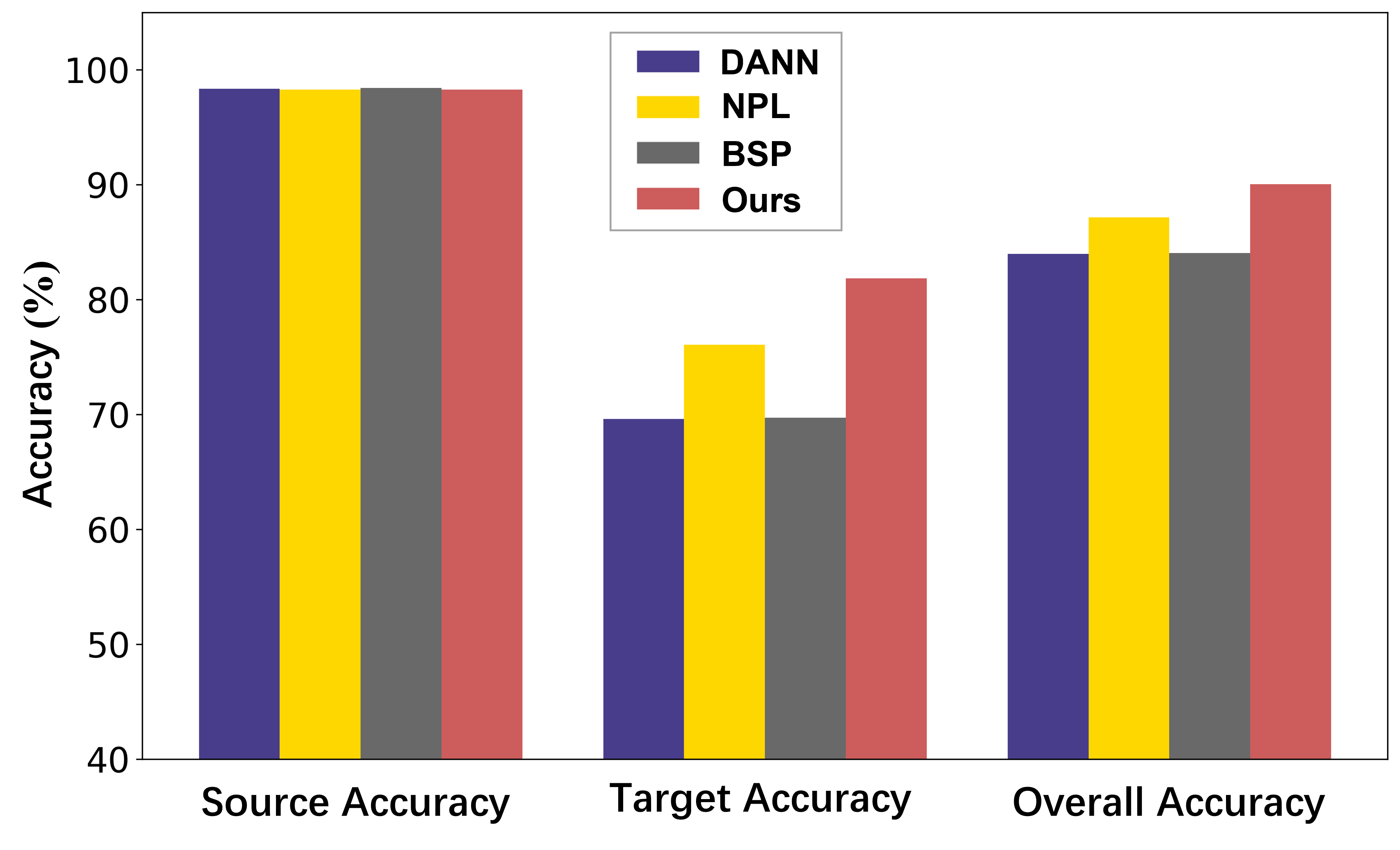} \\
    (a) & (b) & (c) \\
\end{tabular}
}
\caption{
Model analysis on the C $\rightarrow$ R task on the OfficeHome dataset under the UDA setting.
(a)-Coefficient Analysis. We use \textcolor{orange}{orange} for the coefficient of \textbf{$\mathcal{L}_{nap}$} and \textcolor{blue}{blue} for the coefficient of \textbf{$\mathcal{L}_{gsa}$}.
(b)-Distribution Discrepancy. The $\mathcal{A}$-distance measures the distribution discrepancy. The lower $\mathcal{A}$-distance indicates better domain-invariant features.
(c)-Transferability and Discriminability. The higher target accuracy demonstrates that our method enhances transferability while still keeping a strong discriminability.
}
\label{fig:combined_1}
\end{figure*}

\subsection{Model Analysis} \label{sec:exp_model}
\vpara{Ablation Study.}
To verify the effectiveness of each components of \ourloss{} and \ours{},
We designed several variants and compared their classification accuracy on the OfficeHome dataset under the UDA setting.
As shown in Table \ref{tab:ablation},
the version of \ourloss{} w/ $\mathcal{L}_{nap}$, $\mathcal{L}_{gsa}$ 
is based on CDAN \cite{long2018conditional}.
The difference between these variants is whether they utilize the loss of pseudo-labeling and graph spectral penalty or not.
The results illustrate that both pseudo-labeling and graph spectral penalty can improve the performance of the baseline,
and combining these two losses yields better outcomes, which is comparable to cutting-edge methods.
Furthermore, \ours{} with data augmentation and consistency achieves better experimental results, showing the high performance of our overall pipeline.

\vpara{Coefficient Analysis.}
To verify the stability of \ours{}, we further analyze the coefficients of $\mathcal{L}_{nap}$ and $\mathcal{L}_{gsa}$.
As shown in Figure \ref{fig:combined_1}(a),
these results are based on DANN \cite{ganin2015unsupervised}.
Fixing the coefficient of $\mathcal{L}_{nap}$ = 0.2, the coefficient of $\mathcal{L}_{gsa}$ changes from 0.1 to 0.9 with a step size of 0.1.
Fixing the coefficient of $\mathcal{L}_{gsa}$ = 1.0, the coefficient of $\mathcal{L}_{nap}$ changes from 0.1 to 0.9 with a step size of 0.1.
Thus, we can find that different coefficients result in similar results on OfficeHome, showing the insensitivity of \ours{} to these coefficients.

\vpara{Distribution Discrepancy.} 
The $\mathcal{A}$-distance \cite{shaibendavid2007analysis} measures the distribution discrepancy that is defined as  $d_{\mathcal{A}}=2\left(1-2\epsilon\right)$, where $\epsilon$ is the classifier loss to discriminate the source and target domains. 
The smaller $\mathcal{A}$-distance indicates better domain-invariant features.
Figure \ref{fig:combined_1}(b) shows that \ours{} can achieve a lower $d_{\mathcal{A}}$, implying a lower generalization error.

\vpara{Transferability and Discriminability.} 
To examine the trade-off problem with \ours{},
we follow previous work \cite{chen2019transferability} to offer the source accuracy and target accuracy under the UDA setting, as shown in Figure \ref{fig:combined_1}(c).
We can find that various methods achieve similar results of source accuracy, but \ours{} can always achieve higher target accuracy.
This reveals that \ours{} enhances transferability while still keeping a strong discriminability.
This means that \ours{} can find a more suitable utilization of intra-domain and inter-domain information to align target samples properly, jointly balancing the transferability and discriminability as well.

\begin{figure}[!t]
\centering
\resizebox{0.5\textwidth}{!}{
\begin{tabular}{cc}
    \includegraphics{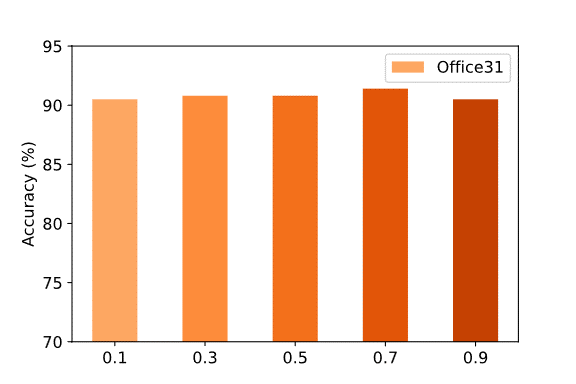} & 
    \includegraphics{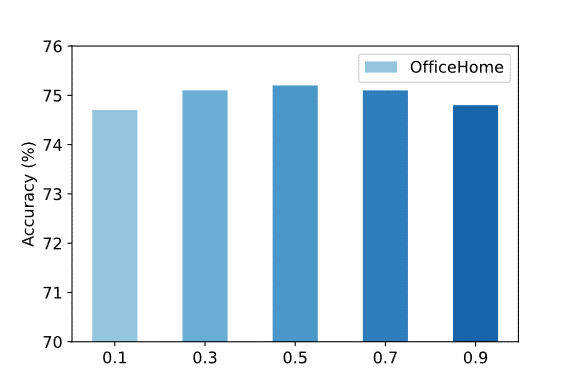} \\
\end{tabular}
}
\caption{
Hyperparameter Sensitivity. 
The hyperparameter $\xi$-exponential moving averaging strategy is used for memory updates.
(a)-The left is on the Office31 dataset.
(b) The right is on the OfficeHome dataset.
These results show that \ours{} is insensitive to this hyperparameter $\xi$.
}
\label{fig:memory}
\end{figure}

\vpara{Hyperparameter Sensitivity.}
To analyze the stability of \ours{},
we design experiments on the hyperparameter $\xi$-exponential moving averaging strategy for memory updates. 
The experimental results of $\xi$ = 0.1, 0.3, 0.5, 0.7, 0.9 are shown in Figure \ref{fig:memory}. 
These results are based on CDAN \cite{long2018conditional}. 
From the series of results,
we can find that in the OfficeHome dataset, the choice of $\xi$ = 0.5 outperforms others.
In addition, the differences between these results are within 0.5\%, 
which means that \ours{} is insensitive to this hyperparameter.

\begin{figure}[!t]
\centering
\resizebox{0.49\textwidth}{!}{
\begin{tabular}{lcccc}
    & UDA & SSDA & MSDA & MTDA \\
    \midrule
    DANN & 
    \includegraphics[width=0.08\textwidth]{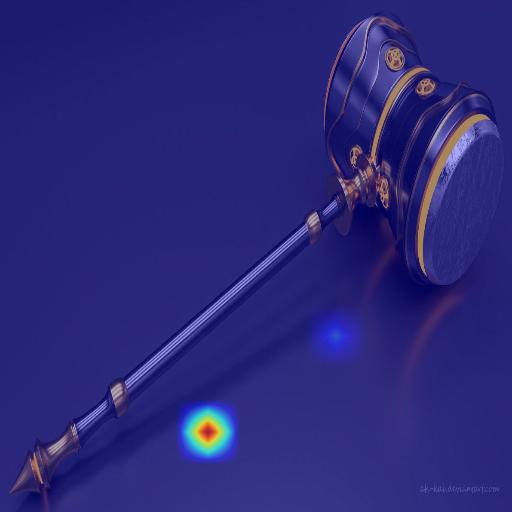} & 
    \includegraphics[width=0.08\textwidth]{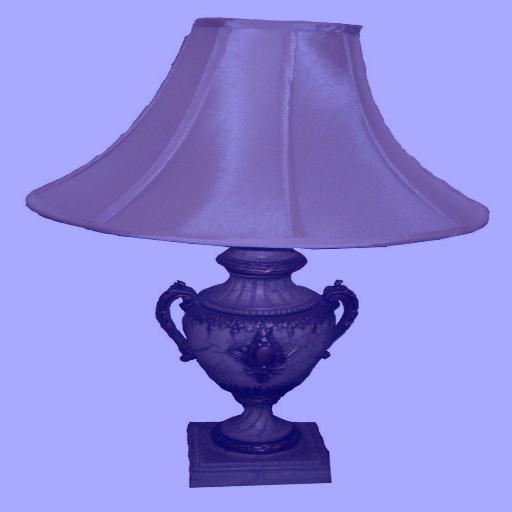} & 
    \includegraphics[width=0.08\textwidth]{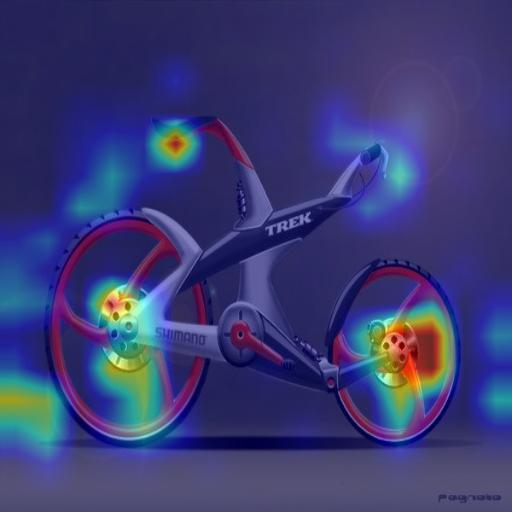} & 
    \includegraphics[width=0.08\textwidth]{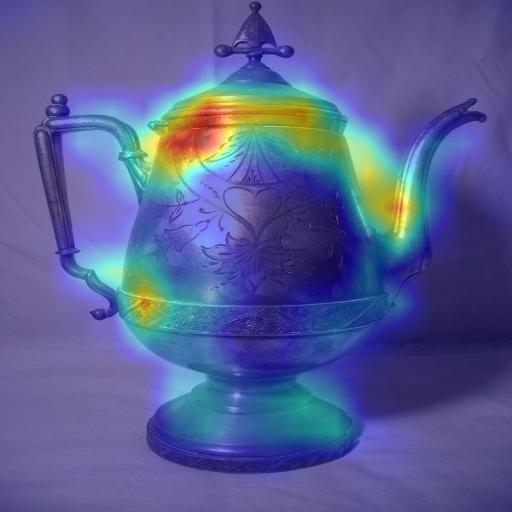} \\
    \midrule
    BSP & 
    \includegraphics[width=0.08\textwidth]{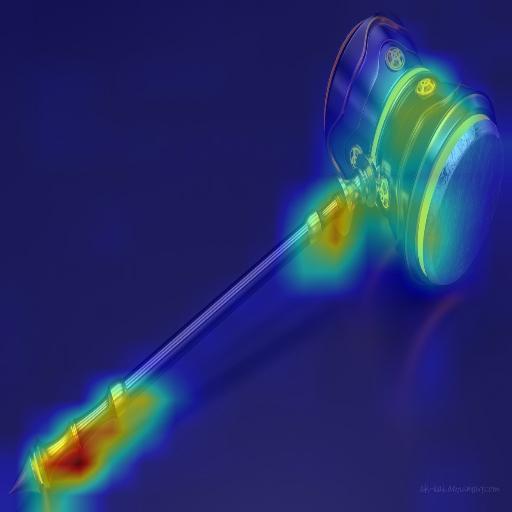} & 
    \includegraphics[width=0.08\textwidth]{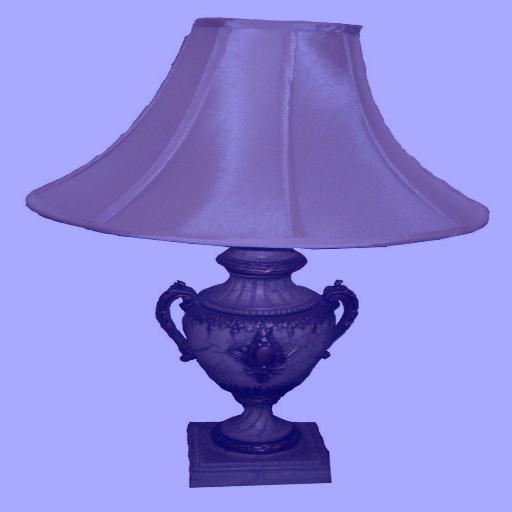} & 
    \includegraphics[width=0.08\textwidth]{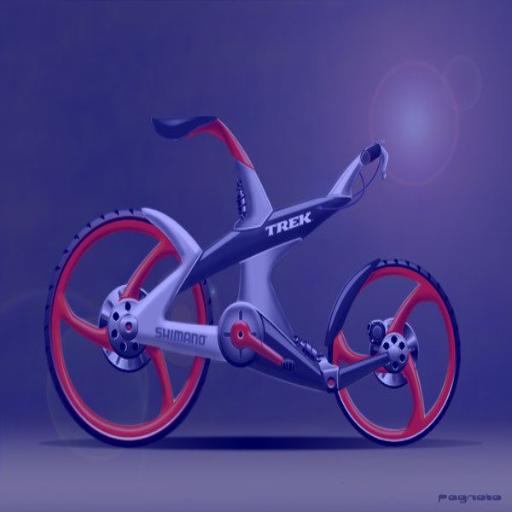} & 
    \includegraphics[width=0.08\textwidth]{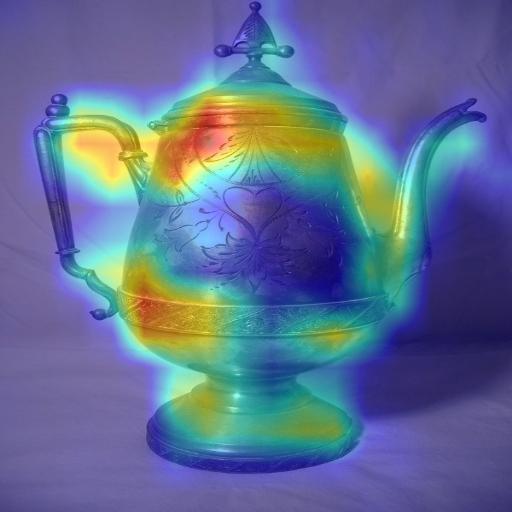} \\
    \midrule
    NPL & 
    \includegraphics[width=0.08\textwidth]{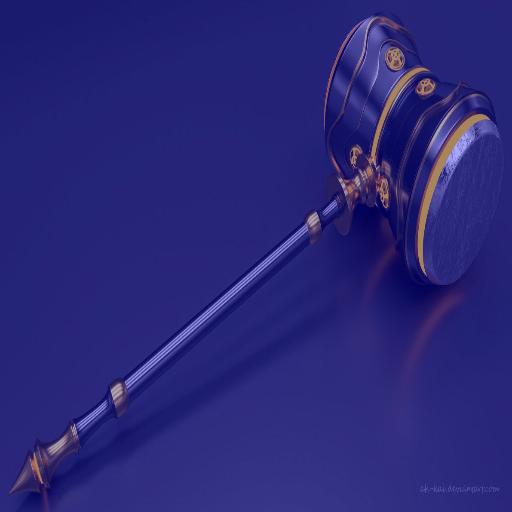} & 
    \includegraphics[width=0.08\textwidth]{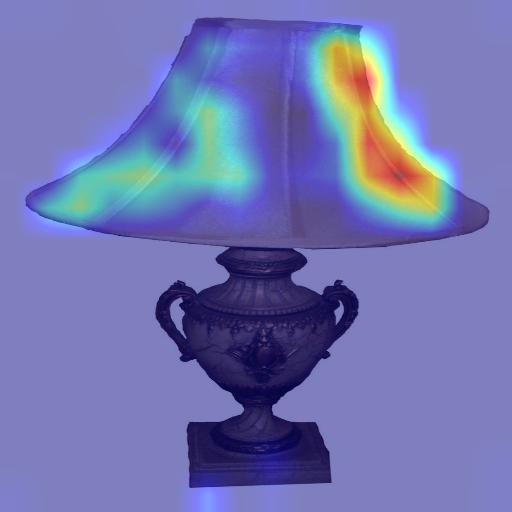} & 
    \includegraphics[width=0.08\textwidth]{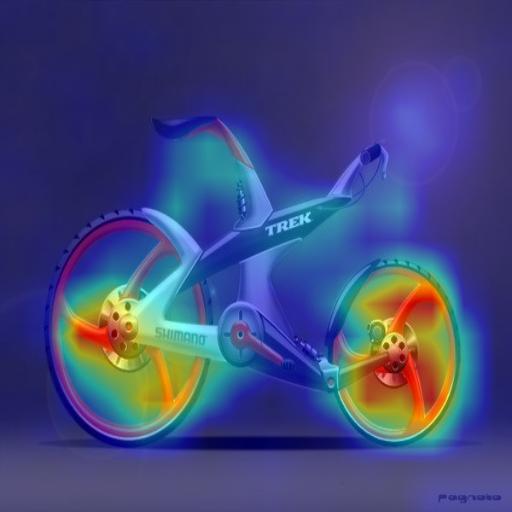} & 
    \includegraphics[width=0.08\textwidth]{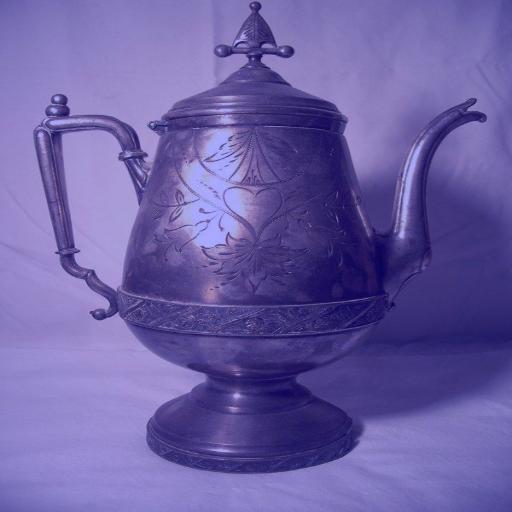} \\
    \midrule
    \ours{} & 
    \includegraphics[width=0.08\textwidth]{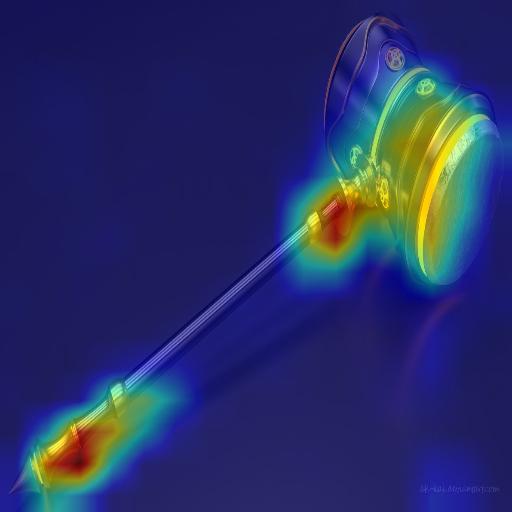} & 
    \includegraphics[width=0.08\textwidth]{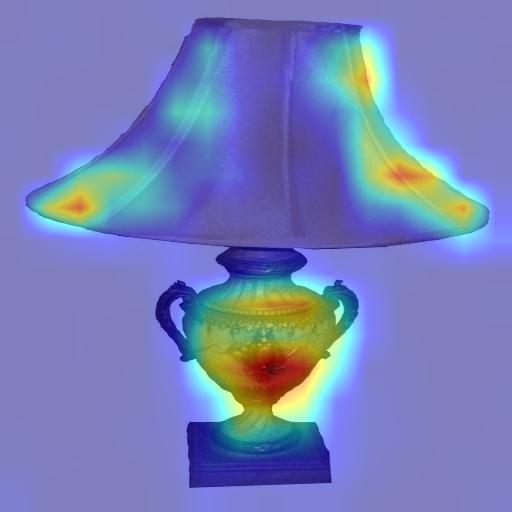} & 
    \includegraphics[width=0.08\textwidth]{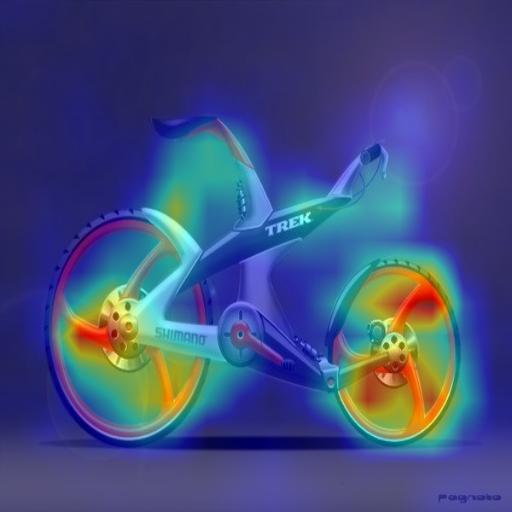} & 
    \includegraphics[width=0.08\textwidth]{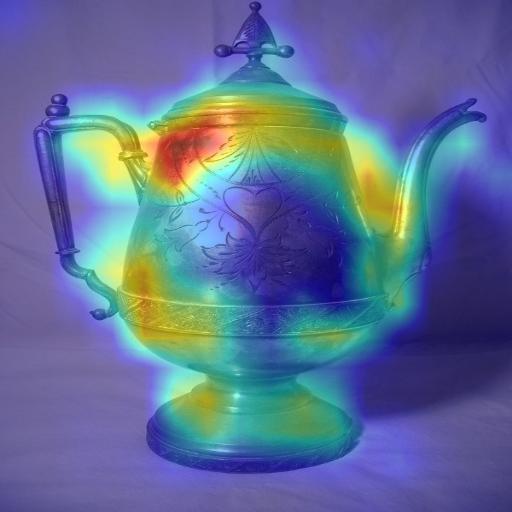} \\
    \bottomrule
\end{tabular}
}
\caption{
Grad-CAM Visualization.
We compare DANN \cite{ganin2015unsupervised}, BSP \cite{cui2020towards}, NPL \cite{lee2013pseudo}, and \ours{} for multiple adaptation setups.
Our method is capable of capturing the crucial regions in images.
For different adaptation setups, its behavior shows consistently strong performance.
}
\label{fig:grad_cam}
\end{figure}

\vpara{Grad-CAM Visualization.} 
We utilize Grad-CAM visualizations to highlight the image regions deemed important by the model. Specifically, we compare the attention maps of DANN \cite{ganin2015unsupervised}, BSP \cite{cui2020towards}, NPL \cite{lee2013pseudo}, and \ours{} across versatile domain adaptation (DA) setups.
DANN is a representative adversarial DA method, BSP adopts matrix decomposition for domain alignment, and NPL employs high-confidence pseudo-labels to guide learning on unlabeled data. These methods are selected as baselines due to their representative characteristics in the DA literature.
As illustrated in Figure~\ref{fig:grad_cam}, \ours{} consistently attends to the complete object, whereas the baseline methods—while effective in their designated DA scenarios—struggle to maintain attention on the primary object when applied to other setups.
This indicates that \ours{} possesses a stronger ability to localize semantically relevant regions that are critical for image classification.
In multi-domain scenarios where a baseline method is best suited, \ours{} produces similar attention patterns, suggesting that it inherits the strengths of specialized approaches.
This still reveals that \ours{} successfully identifies key regions in target domain images across diverse DA setups, showcasing its robustness and versatility.

\begin{figure*}[!t]
\centering
\resizebox{\textwidth}{!}{
\begin{tabular}{ccc}
    \includegraphics[width=0.33\textwidth]{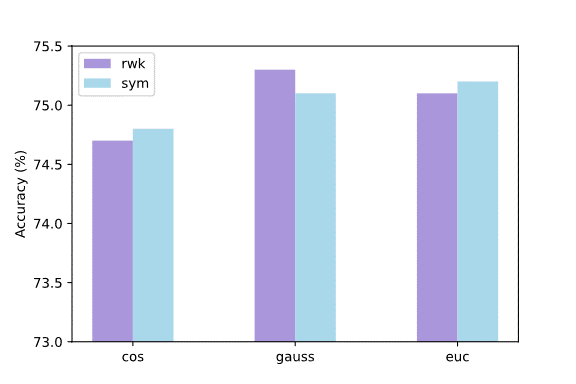} & 
    \includegraphics[width=0.33\textwidth]{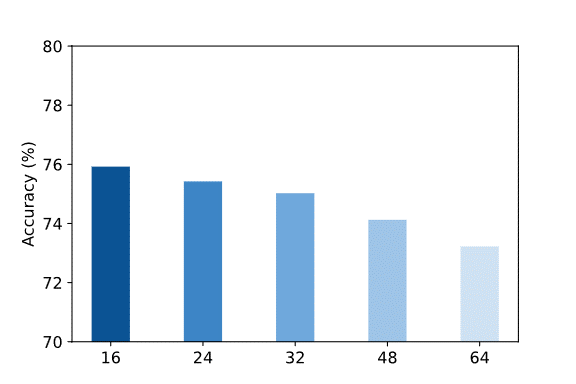} &
        \includegraphics[width=0.33\textwidth]{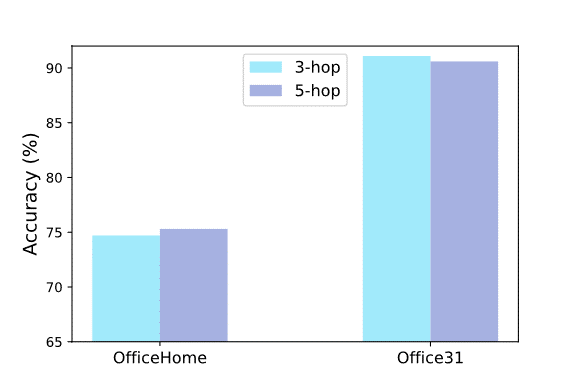} \\
    (a) & (b) & (c) \\
\end{tabular}
}
\caption{
Topology Analysis of graph spectral alignment mechanism in \ours{}.
(a)-Structure Analysis. 
The x-axis: 'cos', 'gauss', and 'euc' indicate the cosine similarity, Gaussian similarity, and Euclidean distance, respectively, during the process of dynamic graph construction, denoted as $\phi(\cdot)$ metric function in Section \ref{sec:graph_construction}.
The y-axis: the accuracy result is the average classification accuracy across 12 tasks on the OfficeHome dataset.
The \textcolor{purple}{purple} shows graph spectral alignment based on $\mathbf{L}_{rwk}$ and the \textcolor{blue}{blue} one based on $\mathbf{L}_{sym}$.
(b)-Order Analysis.
(c)-Density Analysis.
}
\label{fig:combined_2}
\end{figure*}

\vpara{Structure Analysis.}  
To further identify the robustness of \ours{} to different graph structures, we conduct experiments on different types of Laplacian matrices and similarity metrics of graph relations, as illustrated in Figure \ref{fig:combined_2}(a).
We choose two types of commonly-used Laplacian matrix \cite{luxburg2007tutorial}: the random walk Laplacian matrix 
$\mathbf{L}_{rwk} = \mathbf{D}^{-1} \mathbf{A}$, and
the symmetrically normalized Laplacian matrix $\mathbf{L}_{sym} = \mathbf{I} - \mathbf{D}^{-1/2}\mathbf{A}\mathbf{D}^{-1/2}$,
where $\mathbf{D}$ denotes the degree matrix based on the adjacency matrix $\mathbf{A}$.
Here, $\mathbf{L}_{rwk}$ and $\mathbf{L}_{sym}$ achieves comparable performance on OfficeHome dataset. 
Besides, we also tried unnormalized Laplacian matrices during practice, but we found that the performance of the normalized Laplacian matrices is much better than the unnormalized ones.
Furthermore, during the process of graph construction, 
each weighted edge $e_{i j} \in \mathcal{E}_s$ is formulated as a relation between a pair of entities $\phi( \mathbf{f}_i^s, \mathbf{f}_j^s )$, 
where $\phi(\cdot)$ denotes a type of metric function.
The similarity metric function $\phi(\cdot)$ is chosen from cosine similarity, Gaussian similarity, and Euclidean distance.
As for the similarity metric,  the Gaussian similarity brings better performance than the cosine similarity while achieving comparable results with Euclidean distance.
For all these aforementioned experiment results, the differences between them are within 1\% around,  confirming the robustness of the graph spectral alignment mechanism.

\vpara{Order Analysis.}         
To further verify the robustness of \ours{} across different graph scales,
we conduct experiments with varying batch sizes, ranging from 16 to 64 with a step size of 8.
As shown in Figure~\ref{fig:combined_2}(b), smaller batch sizes tend to yield better performance.
This is expected, as when the batch size becomes sufficiently small, the contrastive learning process essentially becomes a sample-to-sample 1-vs-1 alignment, leading to a more fine-grained feature matching.
However, such settings incur significantly higher computational costs on large-scale datasets, making them impractical in real-world applications.
Therefore, a trade-off is needed between alignment granularity and training efficiency.

\vpara{Density Analysis.} 
We also conduct experiments to analyze the impact of different $k$ values in the $k$-hop graph construction, which reflects the density of the dynamic graphs.
As shown in Figure~\ref{fig:combined_2}(c), we compare $k = 3$ and $k = 5$.
This experiment is conducted on both the OfficeHome and Office31 datasets.
The results show that 5-hop graphs outperform 3-hop graphs, particularly on the OfficeHome dataset.
This observation is intuitive—larger $k$ provides more contextual information to each central node, reducing information loss across the graph and enabling a more accurate alignment of distributions.

\begin{figure*}[!t]
\centering
\resizebox{\textwidth}{!}{
\begin{tabular}{cccc}
    \includegraphics[width=0.25\textwidth]{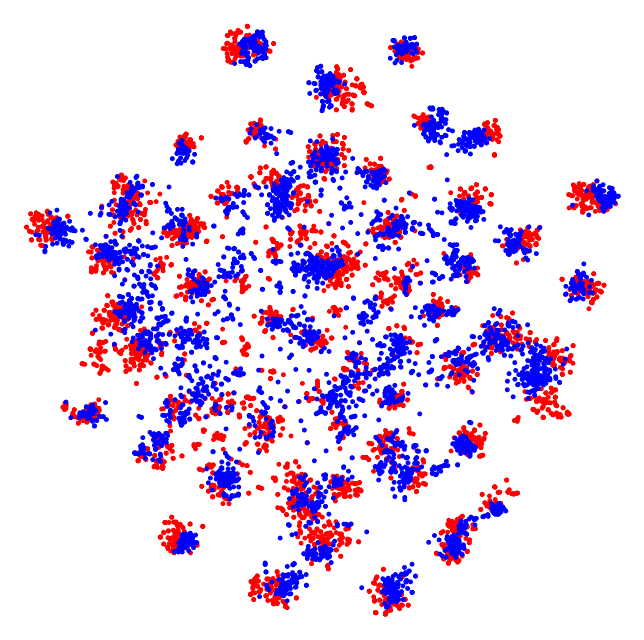} &  
    \includegraphics[width=0.25\textwidth]{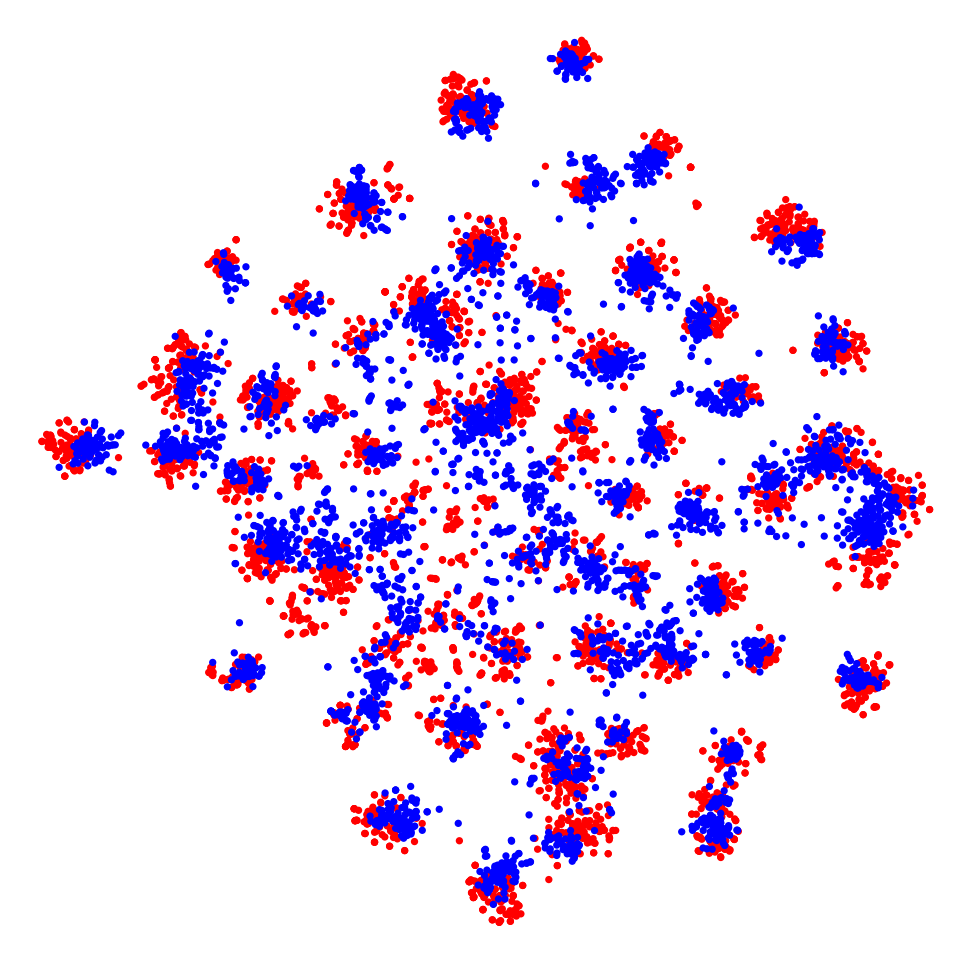} &
    \includegraphics[width=0.25\textwidth]{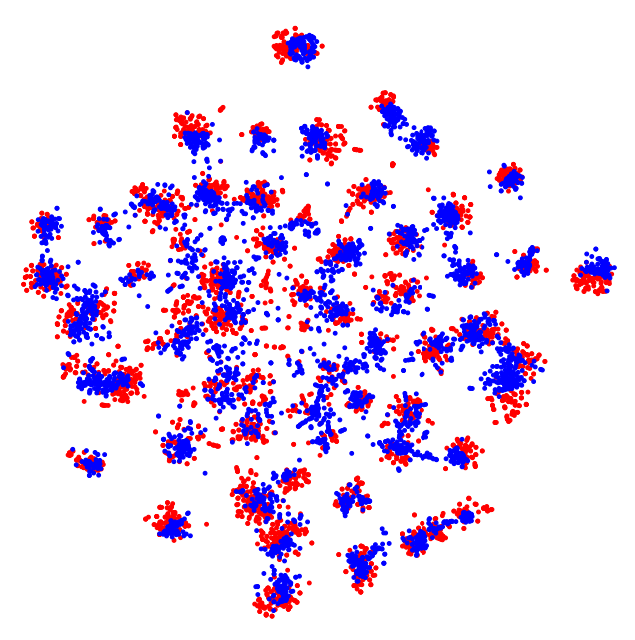} & 
    \includegraphics[width=0.25\textwidth]{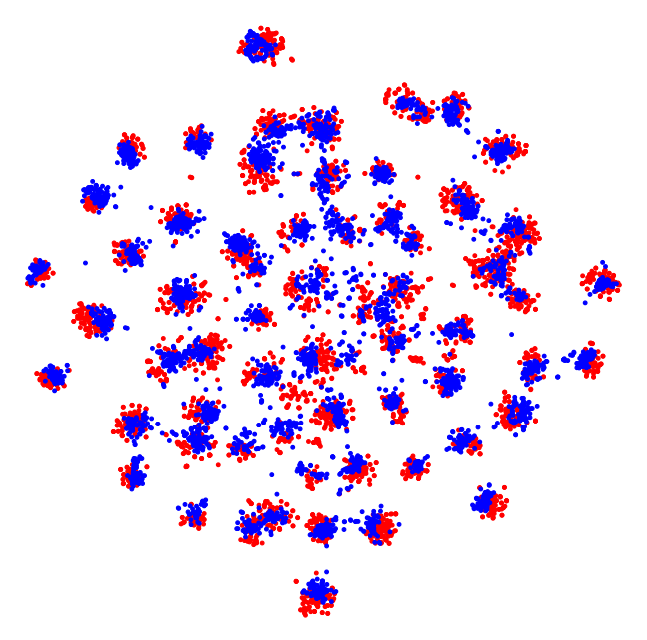} \\
    DANN & BSP & NPL & \ourloss{} \\
    \includegraphics[width=0.25\textwidth]{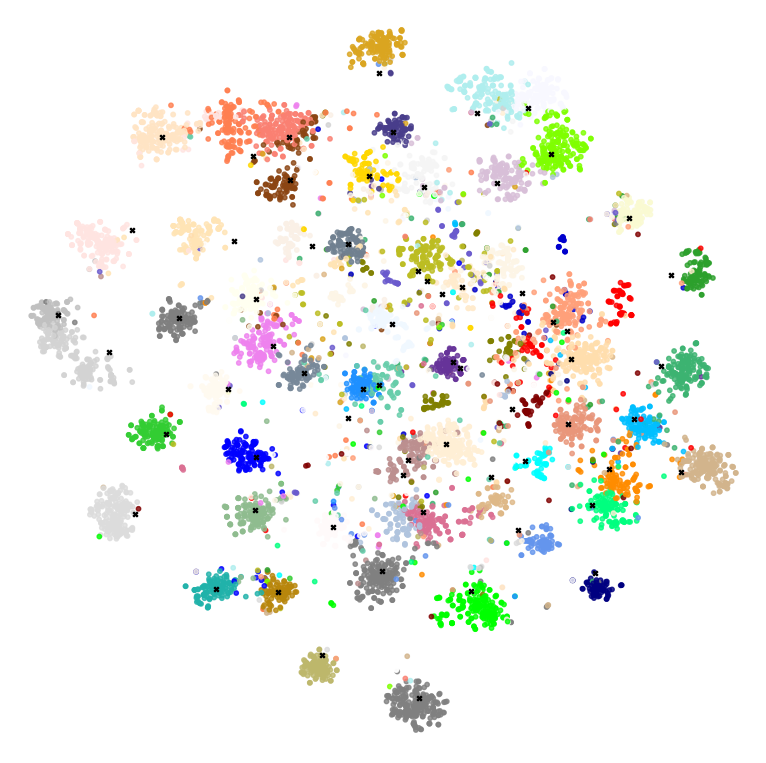} & 
    \includegraphics[width=0.25\textwidth]{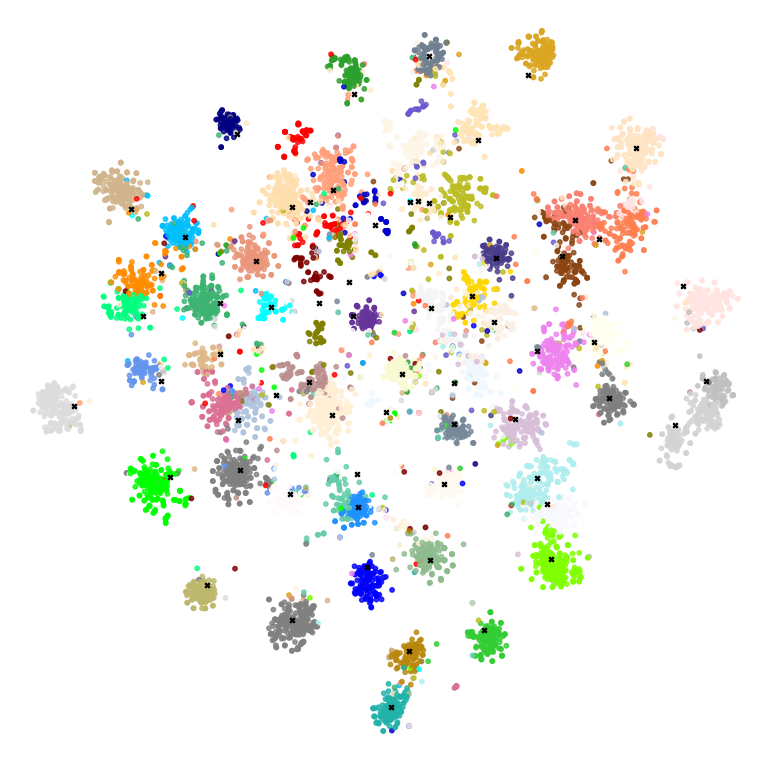} &
    \includegraphics[width=0.25\textwidth]{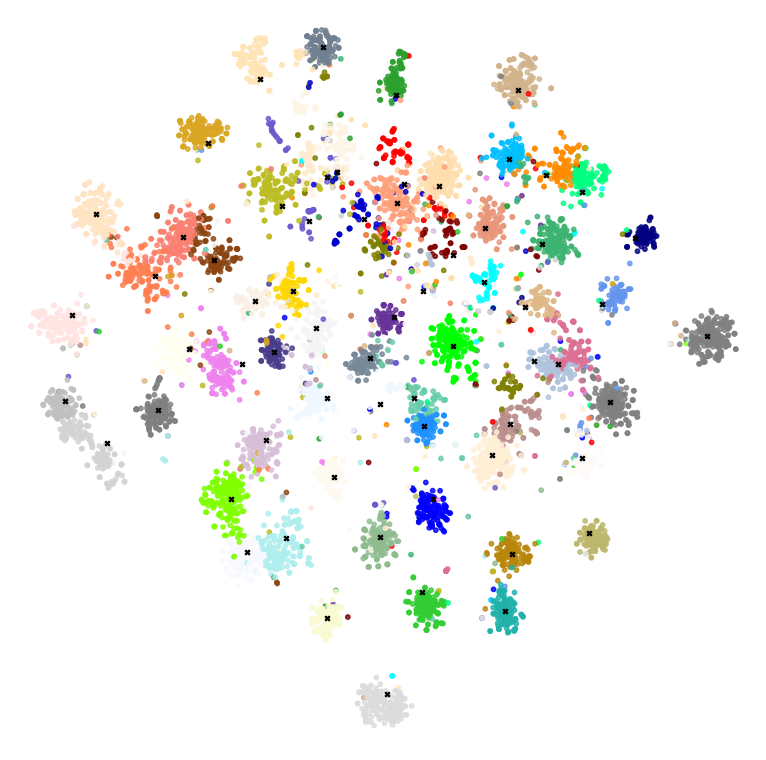} &
    \includegraphics[width=0.25\textwidth]{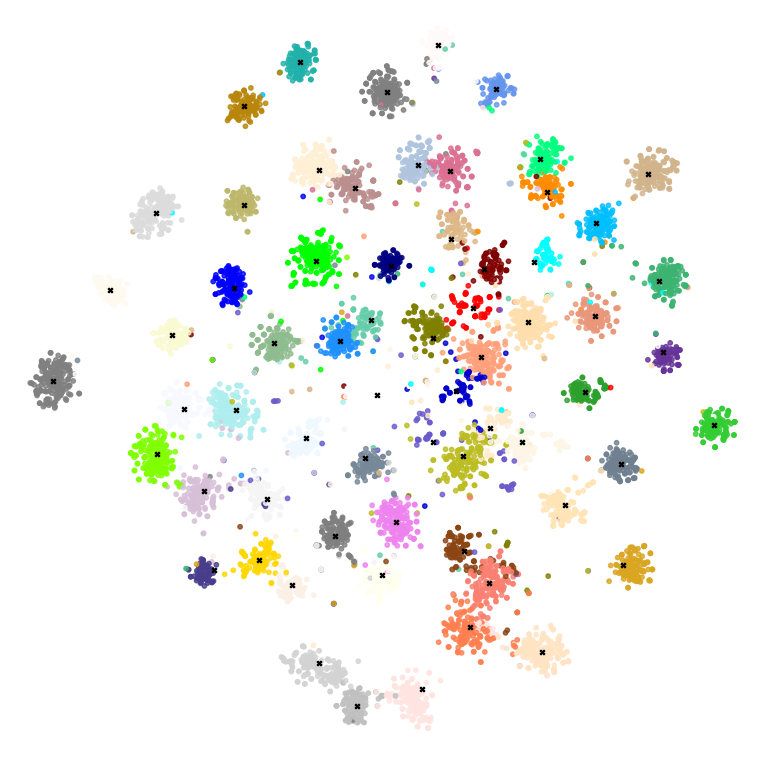} \\
    DANN & BSP & NPL & \ours{} \\
\end{tabular}
}
\caption{
Feature Visualization. 
The t-SNE plot of DANN \cite{ganin2015unsupervised}, BSP \cite{cui2020towards}, NPL \cite{lee2013pseudo}, \ourloss{} and \ours{} features on the OfficeHome dataset in the C $\rightarrow$ R setting.
(Top): We use \textcolor{red}{red} markers for source domain features and \textcolor{blue}{blue} markers for target domain features. 
(Bottom): We plot the class prototypes with black crosses and features with colored circles. 
We observe that \ourloss{} and \ours{} have more tight clusters, showing their transferability and discriminability.
}
\label{fig:tsne_combined}
\end{figure*}

\vpara{Feature Visualization.}
To demonstrate the learning ability of \ours{},
we visualize the features of DANN \cite{ganin2015unsupervised}, BSP \cite{chen2019transferability}, NPL \cite{lee2013pseudo} and \ours{} with the t-SNE embedding \cite{maaten2008visualizing} under the C $\rightarrow$ R setting of OfficeHome Dataset.
DANN \cite{ganin2015unsupervised} is the most classic adversarial DA method.
BSP is a classic domain adaptation method that uses matrix decomposition, and NPL is a classic pseudo-labeling method that uses high-confidence predictions on unlabeled data as true labels.
Thus, we choose them as baselines to compare, as shown in Figure \ref{fig:tsne_combined}.
Here we make some important observations:
(i)-we observe that red markers are closer and overlapping with blue markers of \ourloss{},
which suggests that the source features and target features learned by \ourloss{} are transferred better.
(ii)-we find that the clusters of features of \ours{} are more compact, or say less diffuse than others, 
which suggests that the features learned by \ours{} can attain more desirable discriminative properties.
According to all the figures, the observations are consistent in that the source features and target features learned by \ourloss{} and \ours{} are transferred better.
These observations imply the superiority of \ours{} over discriminability and transferability in unsupervised domain adaptation scenarios.

\section{Conclusion}
We presented SPA++, a unified and versatile framework for domain adaptation based on graph spectral alignment and neighbor-aware propagation. Unlike prior graph-based methods that rely on point-wise matching, SPA++ leverages spectral properties to enable coarse structural alignment and refines target predictions through localized propagation. By integrating augmentation-based consistency regularization, our method extends to broader DA settings such as SSDA, MSDA, and MTDA, and handles complex distribution challenges such as long-tail and subpopulation shifts.
Compared to our earlier conference version, this journal work offers: (1) expanded applicability to versatile DA scenarios; (2) an improved pipeline with stabilized pseudo-labeling via augmentation consistency; (3) theoretical analysis on generalization bounds and spectral smoothing; and (4) extensive empirical validation across multiple benchmarks. Together, SPA++ provides a principled and scalable solution for structure-aware domain adaptation in real-world settings.
Currently, SPA++ is designed for image classification, but its graph and spectral ideas could also be interesting to explore for other modalities such as audio, text, and multi-modal data, or other transfer tasks such as domain generalization. 
Besides, more sophisticated adaptations of our framework for structured prediction tasks such as object detection and semantic segmentation are expected in the future.
We believe deeper integration of graph data mining methods and spectral theory can further enrich both the DA and vision communities.

\begin{acknowledgement}
We gratefully acknowledge Prof. Feng, Dr. Huang, and Dr. Jin for their invaluable contributions to the conference version of this work, which laid an essential foundation for the present journal extension.
This paper is supported by the National Key Research and Development Program of China (No. 2022YFB3304100). It is also partially supported by NSFC under Grants (No. U24A201401), and the Fundamental Research Funds for the Central Universities (226-2024-00049), as well as  (226-2024-00145).
\end{acknowledgement}

\section*{Appendix} 

In this section, we first give a summary of all notations in this paper as Table \ref{tab:notations}.
Then, we present the theoretical foundation and provide proof for the proposed theorem.

\begin{table}[ht]
\caption{Summary of notations by each section.}
\label{tab:notations}
\centering
\resizebox{0.5\textwidth}{!}{
\begin{tabular}{ll}
\toprule
\textbf{Symbol} & \textbf{Meaning} \\
\midrule
\multicolumn{2}{l}{\textit{\ref{sec:graph_construction} Dynamic Graph Construction}} \\
$\mathbf{x}_s, \; \mathbf{x}_t$ & source and target domain samples \\
$\mathbf{f}_s, \; \mathbf{f}_t$ & learned source and target features \\
$\mathcal{G}_s = (\mathcal{V}_s, \mathcal{E}_s)  $ & source domain graph with vertex set $\mathcal{V}_{s}$ and edge set $\mathcal{E}_{s}$ \\
$\mathcal{G}_t = (\mathcal{V}_t, \mathcal{E}_t)$ & target domain graph \\
$\mathbf{A}_s,\; \mathbf{A}_t$ & adjacency matrix of $\mathcal{G}_s$ and $\mathcal{G}_t$ \\
$\delta(\cdot)$ & metric function of edge weight \\
$\theta$ & model trainable parameters \\
\midrule
\multicolumn{2}{l}{\textit{\ref{sec:spectral} Graph Spectral Alignment}} \\
$\Delta$ & graph Laplacian operator \\
$\mathbf{L}_s,\; \mathbf{L}_t$ & Laplacian matrix of $\mathcal{G}_s$ and $\mathcal{G}_t$, respectively \\
$\mathbf{\Lambda}_s,\; \mathbf{\Lambda}_t$ & spectrum of $\mathcal{G}_s$ and $\mathcal{G}_t$ (descending order) \\
$\sigma(\mathcal{G}_s, \mathcal{G}_t)$ & spectral distance between graphs $\mathcal{G}_s$ and $\mathcal{G}_t$ \\
$\mathcal{L}_{gsa}$ & graph spectral alignment loss \\
\midrule
\multicolumn{2}{l}{\textit{\ref{sec:smooth} Smoothing Consistency Analysis}} \\
$k$ & number of nearest neighbors \\
$N_i$ & set of $k$ nearest labeled neighbors of target sample $\mathbf{x}_i$ \\
$\mathbf{p}_i$ & predicted class probability vector for target sample $\mathbf{x}_i$ \\
$\mathbf{p}_i^m$ & sharpened probability vector stored in memory bank \\
$\mathbf{q}_{i,c}$ & propagated probability for class $c$ of sample $\mathbf{x}_i$ \\
$\hat{\mathbf{q}}_{i,c}$ & normalized propagated probability \\
$\hat{\mathbf{y}}_i$ & pseudo-label predicted for target sample $\mathbf{x}_i$ \\
$\tau$ & temperature hyperparameter for sharpening \\
$\mathcal{L}_{nap}$ & neighbor-aware propagation loss \\
\midrule
\multicolumn{2}{l}{\textit{\ref{sec:approach} Overall Architecture}} \\
$\mathcal{D}_s,\; \mathcal{D}_t$ & source and target domain datasets \\
$N_s,\; N_t$ & number of samples in the source and target domains \\
$C_s,\; C_t$ & number of classes in the source and target domains \\
$F(\cdot)$ & feature extractor \\
$C(\cdot)$ & category classifier \\
$D(\cdot)$ & domain classifier \\
$\alpha$ & scaling factor for $\mathcal{L}_{gsa}$  \\
$\beta$ & scaling factor for $\mathcal{L}_{nap}$  \\
$\mathcal{L}_{cls}$ & supervised classification loss \\
$\mathcal{L}_{adv}$ & domain adversarial loss \\
$\mathcal{L}_{total}$ & overall training objective \\
\bottomrule
\end{tabular}
}
\end{table}

\noindent
\textbf{Proof for Theorem \ref{the:Bound}.}
Let 
$
\epsilon_t (h, h^*) 
= \mathbb{E}_{\mathrm{P} \left(\mathcal{G}_t\right)}
\left(\left\|
h \left( \mathcal{G}_t \right) -  h^* \left( \mathcal{G}_t \right)
\right\|\right)
$ 
and 
$
\epsilon_s (h, h^*) 
= \mathbb{E}_{\mathrm{P} \left( \mathcal{G}_s \right)}
\left(\left\| 
h \left( \mathcal{G}_s \right) - h^* \left(\mathcal{G}_s \right)
\right\|\right)
$. 
Firstly, we introduce the following inequality:
$$
\begin{aligned}
\epsilon_t (h) 
& \leq 
\epsilon_t (h^*)  
+ \epsilon_t \left(h, h^*\right) \\
& 
= \epsilon_t (h^*) 
+ \epsilon_s \left(h, h^*\right) 
+ \epsilon_t \left(h, h^*\right) 
- \epsilon_s \left(h, h^*\right)
\end{aligned}
$$

According to Lemma 1 from \cite{shen2018wasserstein}, we proof the following equation:
$$
\begin{aligned}
\epsilon_t (h) 
& \leq 
  \epsilon_t (h^*) 
+ \epsilon_s \left(h, h^*\right) 
+ \epsilon_t \left(h, h^*\right) 
- \epsilon_s \left(h, h^*\right) \\
& \leq 
  \epsilon_t (h^*) 
+ \epsilon_s \left(h, h^*\right) 
+ 2 C_g C_f w( \mathbb{P} (\mathcal{G}_s), \mathbb{P} (\mathcal{G}_t) ) \\
& \leq 
  \epsilon_t (h^*) 
+ \epsilon_s (h)
+ \epsilon_s (h^*) 
+ 2 C_g C_f w( \mathbb{P} (\mathcal{G}_s), \mathbb{P} (\mathcal{G}_t) ) \\
& = 
  \epsilon_s (h) 
+ \eta
+ 2 C_g C_f w( \mathbb{P} (\mathcal{G}_s), \mathbb{P} (\mathcal{G}_t) ) 
\end{aligned}
$$

Then, we link the bound with the empirical risk and labeled sample size by showing, with probability at least $1-\delta$ that:
$$
\begin{aligned}
\epsilon_t (h) 
& \leq 
\epsilon_s (h)
+ 2 C_g C_f w( \mathbb{P} (\mathcal{G}_s), \mathbb{P} (\mathcal{G}_t) ) 
+ \eta \\
& \leq 
\hat{\epsilon}_s (h)
+ \sqrt{\frac{2 d}{N_s} \log \left(\frac{e N_s}{d}\right)  }
+ \sqrt{\frac{1}{2 N_s} \log \left(\frac{1}{\delta}\right) } \\
&
+ 2 C_g C_f w( \mathbb{P} (\mathcal{G}_s), \mathbb{P} (\mathcal{G}_t) ) 
+ \eta
\end{aligned}
$$

Finally, we can derive the following inequality via Cauchy-Schwarz inequality, and provide the conclusion.
$$
\begin{aligned}
\epsilon_t (h) 
& \leq 
\hat{\epsilon}_s (h) 
+ \sqrt{\frac{2 d}{N_s} \log \left(\frac{e N_s}{d}\right)} 
+ \sqrt{\frac{1}{2 N_s} \log \left(\frac{1}{\delta}\right)} \\
&
+ 2 C_g C_f w( \mathbb{P} (\mathcal{G}_s), \mathbb{P} (\mathcal{G}_t) ) 
+ \eta \\
& \leq 
\hat{\epsilon}_s (h) 
+ 
\sqrt{2} 
\sqrt{
\frac{2 d}{ N_s } \log \left(\frac{e N_s}{d}\right)
+ \frac{1}{2 N_s} \log \left(\frac{1}{\delta}\right)
} \\
& 
+ 2 C_g C_f w( \mathbb{P} (\mathcal{G}_s), \mathbb{P} (\mathcal{G}_t) )
+ \eta \\
& = 
\hat{\epsilon}_s (h)
+ \sqrt{
\frac{4 d}{N_s} \log \left(\frac{e N_s}{d}\right)
+ \frac{1}{N_s} \log \left(\frac{1}{\delta}\right)
} \\
&
+ 2 C_g C_f w( \mathbb{P} (\mathcal{G}_s), \mathbb{P} (\mathcal{G}_t) ) 
+ \eta
\end{aligned}
$$
$\hfill\blacksquare$

\noindent
\textbf{Proof for Theorem \ref{the:Transferability}.}
Let a graph feature extractor $f$ as 
a GCN with $k$ layers and $1$-hop graph filter $\Lambda(\mathbf{L})$.
Here we focus on the central node out of simplicity, and denote the representation of node $i \in [1, n]$ in the final layer of GCN by taking a node-wise perspective: 
$
\mathbf{Z}_i^{k} = \sigma ( 
\sum_{j \in \mathcal{N}_i} \Lambda_{i j} \mathbf{Z}_j^{k-1} \mathbf{W}^{k} 
) 
\in \mathbb{R}^d
$, 
where 
$
\Lambda_{i j} = [\Lambda( \mathbf{L})]_{i j} \in \mathbb{R}
$ 
is the weighted link between node $i$ and $j$,
and 
$
\mathbf{W}^{k} \in \mathbb{R}^{d \times d}
$  
is the weight for the $k$-th layer shared across nodes. 
Let denote
$\mathbf{Z}_i^{\ell} \in \mathbb{R}^d$ similarly for $\ell= 1, \cdots, k-1$, 
and 
$\mathbf{Z}_i^0 = \mathbf{X}_i \in \mathbb{R}^d$ 
as the node feature of the center node $i$. 

With the assumption of GCN,
we focus on the $k$-hop ego-graph $\mathcal{G}_i$ centered at $\mathbf{X}_i$ is needed to compute $\mathbf{Z}_i^{k}$ for any $i=1, \cdots, n$.
Let $\mathbf{L}_{\mathcal{G}_i}$ as the out-degree normalised graph Laplacian of $\mathcal{G}_i$ \emph{w.r.t.} the direction from leaves to the centre node in graph $\mathcal{G}_i$. 
We write the $\boldsymbol{\ell}$-th layer representation as follows:
$
[\mathbf{Z}_i^{\ell}]_{k-\ell+1} = \sigma (
[\Lambda (\mathbf{L}_{\mathcal{G}_i}) ]_{k-\ell+1} 
[\mathbf{Z}_i^{\ell-1}]_{k-\ell+1} \mathbf{W}^{\ell})
$.

Given the activation function $\sigma$ is $\rho_\sigma$-Lipschitz function,
and $\forall i$, 
$\max_{\ell} \| \mathbf{Z}_i^{(\ell)} \| \leq c_{\mathbf{z}} $, 
and 
$\max_{\ell} \| \mathbf{W}^{(\ell)} \| \leq c_{\mathbf{w}} $.  
Then, we have
$$
\begin{aligned}
& \| 
[ \mathbf{Z}_{s(i)}^{\ell} ]_{k-\ell} - [ \mathbf{Z}_{t(j)}^{\ell} ]_{k-\ell} 
\|  \\
& \leq 
\| [ 
    \sigma ( 
    [ \Lambda (\mathbf{L}_{\mathcal{G}_i^s} ) ]_{k-\ell+1} 
    [ \mathbf{Z}_{s(i)}^{\ell-1} ]_{k-\ell+1} 
    \mathbf{W}^{\ell} 
    ) \\ 
    & -
    \sigma ( 
    [ \Lambda (\mathbf{L}_{\mathcal{G}_j^t} ) ]_{k-\ell+1}
    [ \mathbf{Z}_{t(j)}^{\ell-1} ]_{k-\ell+1} 
    \mathbf{W}^{\ell}
    )
]_{k-\ell}
\|  \\
& \leq \rho_\sigma c_{\mathbf{w}} 
\| \Lambda (\mathbf{L}_{\mathcal{G}_i^s} ) \| 
\| [ \mathbf{Z}_{s(i)}^{\ell-1} ]_{k-\ell+1} - [ \mathbf{Z}_{t(j)}^{\ell-1} ]_{k-\ell+1} \| \\
& + 
\rho_\sigma c_{\mathbf{w}} c_{\mathbf{z}} 
\| \Lambda (\mathbf{L}_{\mathcal{G}_i^s}) - \Lambda (\mathbf{L}_{\mathcal{G}_j^t}) \|
\end{aligned}
$$

Since 
$[\Lambda (\mathbf{L}_{\mathcal{G}_i^s} )]_{k-\ell+1}$ 
is the principle submatrix of 
$\Lambda (\mathbf{L}_{\mathcal{G}_i^s} )$. 
We equivalently write the above equation as 
$A_{\ell} \leq a A_{\ell-1}+b$, 
where $a$ and $b$ are the coefficient. 
Then, we have
$$
\begin{aligned}
A_{\ell} 
& \leq a A_{\ell-1} + b \\ 
& \leq a^2 A_{\ell-2} + b(1 + a) 
  \leq a^{\ell} A_0 + \frac{a^{\ell}-1}{a-1} b 
\end{aligned}
$$

Therefore, we have an upper bound for the hidden representation difference between $\mathcal{G}_i^s$ and $\mathcal{G}_j^t$ by substituting coefficients $a$ and $b$,
$$
\begin{aligned}
& \|
[\mathbf{Z}_{s(i)}^{\ell}]_{k-\ell} - [\mathbf{Z}_{t(j)}^{\ell}]_{k-\ell}
\| \\
& \leq
(\rho_\sigma c_{\mathbf{w}})^{\ell}
\| \Lambda ( \mathbf{L}_{\mathcal{G}_j^t} ) \|^{\ell}
\| [\mathbf{X}_i^s] - [\mathbf{X}_i^t ] \| \\
& + 
\frac{ 
( \rho_\sigma c_{\mathbf{w}} )^{\ell} 
\| \Lambda (\mathbf{L}_{\mathcal{G}_i^s}) \|^{\ell} - 1 
}{
\rho_\sigma c_{\mathbf{w}} 
\| \Lambda (\mathbf{L}_{\mathcal{G}_i^s}) \| - 1
} 
\rho_\sigma c_{\mathbf{w}} c_{\mathbf{z}} 
\| \Lambda (\mathbf{L}_{\mathcal{G}_i^s}) - \Lambda (\mathbf{L}_{\mathcal{G}_j^t}) \|
\end{aligned}
$$

We can obtain the upper bound for the center node representation 
$
\| [\mathbf{Z}_{s(i)}^{k} ]_0 - [\mathbf{Z}_{t(j)}^{k} ]_0 \| := 
\| \mathbf{Z}_i^s - \mathbf{Z}_j^t \|
$. 
Assume that the difference between any pair of node features do not exceed a constant $c_{\mathbf{x}}$, 
namely, 
$
\|
[\mathbf{X}_i^s]-[\mathbf{X}_j^t]
\| 
\leq c_{\mathbf{x}}
$.
We have $\forall i$, 
$\| \Lambda (\mathbf{L}_{\mathcal{G}_i^s}) \| \leq c_\mathbf{L}$ out of the normalized graph Laplacians and $\Lambda (\cdot)$ is a linear function for $\mathbf{L}$, we have
$$
\begin{aligned}
\| \mathbf{Z}_i^s - \mathbf{Z}_j^t \|
& \leq (\rho_\sigma c_{\mathbf{w}} c_\mathbf{L})^k c_{\mathbf{x}}  \\
& + \frac{ (\rho_\sigma c_{\mathbf{w}} c_\mathbf{L} )^k - 1 }{
\rho_\sigma c_{\mathbf{w}} c_\mathbf{L} - 1
} 
\rho_\sigma c_{\mathbf{w}} c_\mathbf{z} 
\| \Lambda (\mathbf{L}_{\mathcal{G}_i^s}) - \Lambda (\mathbf{L}_{\mathcal{G}_j^t}) \| \\
& \leq 
C_1 \|\mathbf{L}_{\mathcal{G}_i^s} - \mathbf{L}_{\mathcal{G}_j^t} \| + C_2
\end{aligned}
$$
Let 
$
C_1 = 
\frac{ (\rho_\sigma c_{\mathbf{w}} c_\mathbf{L} )^k - 1 }{
\rho_\sigma c_{\mathbf{w}} c_\mathbf{L} - 1
} 
\rho_\sigma c_{\mathbf{w}} c_\mathbf{z} 
$ and 
$
C_2 = 
(\rho_\sigma c_{\mathbf{w}} c_\mathbf{L})^k c_{\mathbf{x}}
$.
Finally, let $\xi_1=\chi_1$ and $\xi_2=\chi_2$, concluding the proof, 
we can finally conclude the following equation
$$ 
\begin{aligned}
\| f(\mathcal{G}_s) - f(\mathcal{G}_t) \| 
& \leq 
\frac{C_1}{N_s N_t} 
\sum_{i=1}^{N_s} \sum_{j=1}^{N_t}
\| \mathbf{L}_{\mathcal{G}^s_i} - \mathbf{L}_{\mathcal{G}^t_j} \| 
+ C_2 \\
& \leq 
C_1 \Delta (\mathcal{G}_s, \mathcal{G}_t) + C_2
\end{aligned}
$$
$\hfill\blacksquare$

\noindent
\textbf{Proof for Theorem \ref{the:subspace}.}
Following \cite{horn2012matrix}, the squared Frobenius norm of a matrix difference between two graph Laplacians can be bounded by
$
\sum_{i=1}^n (\Lambda_{s(i)} - \Lambda_{t(i)} )^2  
\leq \|\mathbf{L}_s - \mathbf{L}_t \|
$, 
where $\Lambda_s^i$ and $\Lambda_t^i$ are the $i$-th singular value of the graph Laplacians based on the source graph and target graph, respectively, in descending order. 

However, the subspace matrices $\mathbf{L}_s^l$ and $\mathbf{L}_t^l$ are a special case due to the subspace override of the projector 
$\mathbf{M}=\mathbf{U}_t^l \mathbf{U}_s^{l-1}$, 
because
$$
\begin{aligned}
& \| \mathbf{L}_s^l - \mathbf{L}_t^l \| = 
  \| \mathbf{M} \mathbf{U}_s^l 
     \boldsymbol{\Lambda}_s^l - \mathbf{U}_t^l 
     \boldsymbol{\Lambda}_t^l \| = 
  \| \mathbf{U}_t^l \boldsymbol{\Lambda}_s^l - \mathbf{U}_t^l \boldsymbol{\Lambda}_t^l \|  \\
& = \| \mathbf{U}_t^l \boldsymbol{\Lambda}_s^l \| 
  + \| \mathbf{U}_t^l \boldsymbol{\Lambda}_t^l \| 
  - 2 \operatorname{tr}
  \left(\boldsymbol{\Lambda}_s^{l -1} \mathbf{U}_t^l \mathbf{U}_t^l \boldsymbol{\Lambda}_t^l\right) \\
& = \|\boldsymbol{\Lambda}_s^l \| 
+ \|\boldsymbol{\Lambda}_t^l \|
- 2 \operatorname{tr} ( \boldsymbol{\Lambda}_s^{l -1} \boldsymbol{\Lambda}_t^l ) \\
& = \sum_{i=1}^l ( \Lambda_s^{i} )^2
+ \sum_{i=1}^l ( \Lambda_t^{i} )^2
- 2 \sum_{i=1}^l ( \Lambda_s^{i} \cdot \Lambda_t^{i} ) 
= \sum_{i=1}^l ( \Lambda_s^{i} - \Lambda_t^{i} )^2
\end{aligned}
$$

Therefore, we can directly compute the Frobenius inner product of the diagonal matrices $\boldsymbol{\Lambda}_s^l$ and $\boldsymbol{\Lambda}_t^l$, 
which is simply the sum of the product of the singular values. 
Consequently follows for $l + 1$ 
and $( \Lambda_s^{l+1} - \Lambda_t^{l+1} )^2 \neq 0$.
For $1 < l < n $, we have: 
$$
\begin{aligned}
\|\mathbf{L}_s^l - \mathbf{L}_t^l \|
& < \sum_{i=1}^{l+1} 
(\Lambda_s^i - \Lambda_t^i )^2 \\
& < \sum_{i=1}^n
(\Lambda_s^i - \Lambda_t^i )^2 
\leq 
\| \mathbf{L}_s - \mathbf{L}_t \|
\end{aligned}
$$.
$\hfill\blacksquare$

\noindent
\textbf{Proof for Theorem \ref{the:lpa}.}
Let $\bar{\mathbf{A}}_{i j} = \mathbf{A}_{i j} / \mathbf{D}_{ii} $ denoted as the normalized edge weight between node $i$ and $j$.
It is clear that $\sum_{j \in \mathcal{N}_i} \bar{\mathcal{A}}_{i j} = 1$. 
Given that $\mathcal{M}$ is differentiable, 
we perform a first-order Taylor expansion with Peano's form of remainder at $\mathbf{x}_i$ for $\sum_{j \in \mathcal{N}_i } \bar{\mathbf{A}}_{i j} \mathbf{y}_j $:
$$
\begin{aligned}
& 
\sum_{j \in \mathcal{N}_i} \bar{\mathbf{A}}_{i j} \mathbf{y}_j = 
\sum_{j \in \mathcal{N}_i} \bar{\mathbf{A}}_{i j}  \mathcal{M} (\mathbf{x}_j ) \\
& = 
\sum_{j \in \mathcal{N}_i} \bar{\mathbf{A}}_{i j} 
\left(
    \mathcal{M} (\mathbf{x}_i) + 
    \frac{ \partial \mathcal{M} (\mathbf{x}_i) }{ \partial \mathbf{x}^{-1} } 
    (\mathbf{x}_j-\mathbf{x}_i) + 
    o( \| \mathbf{x}_j - \mathbf{x}_i \| )
\right) \\
& = \mathcal{M} (\mathbf{x}_i) 
+ \frac{\partial \mathcal{M}(\mathbf{x}_i)}{\partial \mathbf{x}^{-1}} 
\sum_{j \in \mathcal{N}_i} \bar{\mathbf{A}}_{i j} (\mathbf{x}_j-\mathbf{x}_i)
\\
& + \sum_{j \in \mathcal{N}_i} \bar{\mathbf{A}}_{i j} o( \| \mathbf{x}_j - \mathbf{x}_i \| ) \\
& = y_i 
- \frac{\partial \mathcal{M} (\mathbf{x}_i)}{\partial \mathbf{x}^{-1}} 
\epsilon_i
+ \sum_{j \in \mathcal{N}_i} \bar{\mathbf{A}}_{i j} o( \| \mathbf{x}_j - \mathbf{x}_i \|)
\end{aligned}
$$

According to the Cauchy-Schwarz inequality and the $C_m$-Lipschitz property, we have
$$
\left| 
\frac{\partial \mathcal{M} (\mathbf{x}_i)}{\partial \mathbf{x}^{-1}} \epsilon_i
\right| 
\leq
\left\| 
\frac{\partial \mathcal{M} (\mathbf{x}_i)}{\partial \mathbf{x}^{-1}} 
\right\|
\| \epsilon_i \| 
\leq 
C_m \| \epsilon_i \| 
$$

Therefore, the approximation of $\mathbf{y}_i$ is bounded by
$$
\begin{aligned}
& | \mathbf{y}_i - \textstyle \sum_{j \in \mathcal{N}_i} \bar{\mathbf{A}}_{i j} \mathbf{y}_j | \\
= & 
\left| \textstyle 
\frac{\partial \mathcal{M} (\mathbf{x}_i)}{\partial \mathbf{x}^{-1}} \epsilon_i
- \sum_{j \in \mathcal{N}(i)} \bar{\mathbf{A}}_{i j} 
o( \| \mathbf{x}_j - \mathbf{x}_i \| )
\right| \\
\leq & 
\left| \textstyle 
\frac{\partial \mathcal{M} (\mathbf{x}_i)}{\partial \mathbf{x}^{-1}} \epsilon_i
\right|
+
\left|
\sum_{j \in \mathcal{N}_i} \bar{\mathbf{A}}_{i j}  
o( \| \mathbf{x}_j - \mathbf{x}_i \| )
\right| \\
\leq & 
C_m \left\|\epsilon_i\right\|_2
+o(\textstyle \max_{j \in \mathcal{N}_i}(\| \mathbf{x}_j - \mathbf{x}_i \|))
\end{aligned}
$$
$\hfill\blacksquare$

\bibliographystyle{fcs}
\bibliography{reference}

\begin{biography}{figures_info/xzq}
Zhiqing Xiao received a BS degree from Zhejiang University, China, in 2020. She is currently working toward a PhD degree at the College of Computer Science and Technology, Zhejiang University, China. Her main research interests include knowledge transfer, vision-language models, and graph data mining.
\end{biography}

\begin{biography}{figures_info/whb}
Haobo Wang received his PhD degree in computer science from Zhejiang University, China, in 2023. He is now an assistant professor at the School of Software Technology, Zhejiang University, China. His main research interests include machine learning and data mining, especially on weakly-supervised learning and large language models.
\end{biography}

\begin{biography}{figures_info/lx}
Xu Lu is a senior undergraduate student at the College of Computer Science and Electronic Engineering, Hunan University, pursuing a degree in Software Engineering. His academic focus lies in deep learning and natural language processing, particularly in domain generalization and time series prediction.
\end{biography}

\begin{biography}{figures_info/ywt}
Wentao Ye received a BS degree from Zhejiang University, China, in 2023. He is currently working toward a PhD degree at the College of Computer Science and Technology, Zhejiang University, China. His main research interests include natural language processing, large language models, and AI safety.
\end{biography}

\begin{biography}{figures_info/cg}
Gang Chen (Member of IEEE, ACM) received his PhD degree in computer science from Zhejiang University, China, in 1998. He is now a professor at the College of Computer Science and Technology, Zhejiang University, China. His main research interests include database management technology, intelligent computing based on big data, and massive internet systems. He is a member of IEEE, ACM, standing member of the CCF Database Professional Committee.
\end{biography}

\begin{biography}{figures_info/zjb}
Junbo Zhao received his PhD degree in computer science from New York University, United States in 2019. He is now an assistant professor at the College of Computer Science and Technology, Zhejiang University, China. His research interests include large language models, table pre-training, machine learning, and AI+X.
\end{biography}

\end{document}